\documentclass{svproc}
\usepackage[a4paper]{geometry}
\usepackage{booktabs}
\usepackage{graphicx}
\usepackage{placeins}
\usepackage{amsmath}
\usepackage{amssymb}
\usepackage{float}
\usepackage{multirow}
\usepackage{tabularx}
\usepackage[version=4]{mhchem}
\usepackage{subcaption}
\usepackage{hyperref}
\usepackage{titling}
\usepackage{authblk}
\usepackage{array}
\usepackage{tikz}
\usepackage[inline]{enumitem}
\usepackage{yhmath}
\usepackage[version=4]{mhchem}
\usepackage{comment}
\usepackage{upgreek,textgreek}
\usepackage{fancyhdr}
\usepackage{subcaption}
\usepackage[colorinlistoftodos]{todonotes}

\makeatletter
\let\Title\@title
\makeatother
\usepackage[T2A]{fontenc}

\usepackage{csquotes}

\usepackage[
    natbib=true,
    style=numeric,
    sorting=none
]{biblatex}
\usepackage{xargs}                      

\newcommandx{\unsure}[2][1=]{\todo[linecolor=red,backgroundcolor=red!25,bordercolor=red,#1]{#2}}
\newcommandx{\change}[2][1=]{\todo[linecolor=blue,backgroundcolor=blue!25,bordercolor=blue,#1]{#2}}
\newcommandx{\info}[2][1=]{\todo[linecolor=green,backgroundcolor=green!25,bordercolor=green,#1]{#2}}
\newcommandx{\improvement}[2][1=]{\todo[linecolor=Plum,backgroundcolor=Plum!25,bordercolor=Plum,#1]{#2}}
\newcommandx{\thiswillnotshow}[2][1=]{\todo[disable,#1]{#2}}
\usetikzlibrary{fadings}
\usetikzlibrary{patterns}
\usetikzlibrary{shadows.blur}
\usetikzlibrary{shapes}
\usepackage{setspace}
\DeclareUnicodeCharacter{2212}{\textendash}
\DeclareUnicodeCharacter{2032}{\textendash}

\mainmatter

\title{\textbf{Cardiac mortality prediction in patients undergoing PCI based on real and synthetic data}}

\author{Daniil Burakov\inst{1} Ivan Petrov\inst{1} Dmitrii Khelimskii\inst{2}\ Ivan Bessonov\inst{3} Mikhail Lazarev\thanks{Corresponding author}}

\affil[1]{HSE University, Moscow, Russian Federation \email{diburakov@hse.ru, mvlazarev@hse.ru, ivalpetrov@edu.hse.ru}}
\affil[2]{Meshalkin National Medical Research Center, Ministry of Health of Russian Federation, Novosibirsk, Russian Federation
\email{dkhelim@mail.ru}}
\affil[3]{Tyumen Cardiology Research Center, Tomsk National Research Medical Center, Russian Academy of Sciences, Tomsk 625026, Russian Federation}

\authorrunning{Daniil Burakov et al.} 
\titlerunning{Mortality prediction}
\tocauthor{Daniil Burakov et al.}
\date{}

\begin{document}
\maketitle

\begin{abstract}

Patient status, angiographic and procedural characteristics encode crucial signals for predicting long-term outcomes after percutaneous coronary intervention (PCI). 
The aim of the study was to develop a predictive model for assessing the risk of cardiac death based on the real and synthetic data of patients undergoing PCI and to identify the factors that have the greatest impact on mortality. We analyzed 2,044 patients, who underwent a PCI for bifurcation lesions. The primary outcome was cardiac death at 3-year follow-up. Several machine learning models were applied to predict three-year mortality after PCI. To address class imbalance and improve the representation of the minority class, an additional 500 synthetic samples were generated and added to the training set. To evaluate the contribution of individual features to model performance, we applied permutation feature importance. An additional experiment was conducted to evaluate how the model’s predictions would change after removing non-informative features from the training and test datasets. Without oversampling, all models achieve high overall accuracy (0.92-0.93), yet they almost completely ignore the minority class. Across models, augmentation consistently increases minority-class recall with minimal loss of AUROC, improves probability quality, and yields more clinically reasonable risk estimates on the constructed severe profiles. According to feature importance analysis, four features emerged as the most influential: Age, Ejection Fraction, Peripheral Artery Disease, and Cerebrovascular Disease. On the external dataset, a decline in the target metric AUC-ROC to the range of 0.59–0.71 is observed across all models.
These results show that straightforward augmentation with realistic and extreme cases can expose, quantify, and reduce brittleness in imbalanced clinical prediction using only tabular records, and motivate routine reporting of probability quality and stress tests alongside headline metrics.

\textbf{Keywords}: machine learning, tabular data, mortality prediction, bifurcation lesions, medical data, uncertainty, cardiac death, PCI, synthetic data.

\end{abstract}

\section{Introduction}

Coronary artery disease remains one of the leading causes of morbidity and mortality worldwide \cite{sawaya2016contemporary, d2017incidence, naganuma2013long}, despite significant advances in modern cardiology and the development of interventional technologies. It is characterized by chronic damage to the coronary arteries by atherosclerotic plaques, which leads to impaired coronary blood flow, myocardial ischemia, and the development of acute coronary syndromes. In recent decades, significant improvements have been achieved in the diagnosis and treatment of CAD, including the widespread introduction of percutaneous coronary interventions (PCI) and the use of modern antiplatelet drugs. However, even with successful intervention, there remains a high risk of adverse long-term outcomes, including restenosis, recurrent myocardial infarction, and cardiovascular mortality.

Predicting long-term outcomes in patients with CAD is a difficult task due to the multifactorial nature of the disease. Outcomes depend not only on the anatomical characteristics of the affected vessels and the features of the intervention performed, but also on a wide range of clinical factors \cite{wu2006risk}, as well as behavioral and genetic predispositions. High variability in the combination of risk factors leads to significant difficulties in constructing universal predictive models.

In recent years, machine learning methods have been increasingly applied in healthcare, offering new tools for processing complex, high-dimensional clinical datasets \cite{esteva2021deep, grasso1995automated}. One of their promising applications lies in predicting long-term outcomes in patients after cardiovascular interventions \cite{benjamins2019primer}. However, this task is associated with significant challenges, including class imbalance in clinical outcomes, data heterogeneity, limited dataset sizes, and difficulties in interpretability of models \cite{holzinger2019causability}. Classical approaches such as logistic regression \cite{hosmer2013applied} or support vector machines \cite{suthaharan2016support} provide interpretable results but often underperform in predictive accuracy. In contrast, ensemble methods such as gradient boosting like CatBoost \cite{prokhorenkova2018catboost} and XGBoost \cite{chen2016xgboost}, or multilayer perceptrons \cite{rosenblatt1957perceptron} demonstrate superior performance in terms of discrimination metrics (e.g., AUC-ROC, F1-score), though at the cost of model transparency.

Synthetic data are increasingly employed in supervised classification to mitigate common practical constraints in clinical research—limited sample size, severe class imbalance, and data-sharing/privacy restrictions—by providing additional training examples that expand the minority class, preserve utility for downstream models, and enable reproducible experimentation without exposing patient records. Generative techniques range from classical oversampling algorithms (e.g., SMOTE \cite{chawla2002smote} and its variants) to deep generative models such as conditional GANs \cite{xu2019modeling} (CTGAN and derivatives), variational autoencoders \cite{kingma2019introduction}, and recently diffusion-based tabular generators \cite{li2025diffusion}; each approach has different trade-offs in fidelity, diversity, and privacy. Empirically, GAN-based methods and certain algorithmic oversamplers often improve classifier sensitivity on rare classes, while some generative methods (particularly naively trained VAEs \cite{xu2019modeling} or poorly tuned copula models) can produce samples that reduce discriminative performance or distort feature relationships if fidelity is insufficient. Diffusion models for tabular data have recently emerged as a promising alternative \cite{kotelnikov2023tabddpm}, showing strong sample quality and improved downstream utility in several benchmarks, but they typically require careful conditioning and more data to train reliably.

Our study aims to develop a predictive model for assessing the risk of cardiac death based on clinical, angiographic, and procedural characteristics of the real and synthetically generated patient undergoing PCI and to identify the factors that have the greatest impact on mortality.

\section{Methods}

The study was conducted based on a multicenter all-comer registry to treat patients with bifurcation lesions of the coronary arteries \parencite{NCT03450577}. A total of 2,044 patients were included in the analysis, who underwent a percutaneous coronary intervention for bifurcation lesions between 2018 and 2020. The primary clinical outcome was cardiac death. Death was considered to have a cardiac cause unless an unequivocal noncardiac cause could be established. All information about the patient, including demographics, medications, laboratory data, angiographic data, procedural data, outcomes was collected using a web-based REDCap system.  Clinical follow-up data were collected at 1 and 3 years after index PCI. Follow-up was performed either via direct phone contact with the patient or during a visit of the patient to the hospital. The study was carried out in accordance with the Helsinki declaration and was approved by the institutional review board. 
The patients were divided into two groups — training (N = 1635, positive class = 127) and test (N = 409, positive class = 31) — using stratified sampling to preserve the class distribution. On the training subset, a 10-fold cross-validation procedure was applied to estimate the average model performance. This approach is particularly beneficial when working with limited datasets, as it reduces the variance of performance estimates and provides a more reliable evaluation of model generalization.

Prior to the initiation of model training, the data underwent a preprocessing procedure. Specifically, features with more than 500 missing entries were removed from the dataset. Missing values in categorical features were replaced with the most frequently occurring category, while continuous features were filled using the IterativeImputer method \cite{van2011mice}. Categorical features were encoded using the One-Hot encoding method.

The analysis of variance (ANOVA) F-value was used as a univariate feature selection criterion. For each feature, the F-statistic compares the variance of the feature between outcome classes to the variance within each class. A higher F-value indicates that the feature provides stronger discrimination with respect to the target variable.

For a given feature $x$ across two outcome classes (positive and negative), 
the ANOVA $F$-statistic is defined as: $F = \frac{MS_B}{MS_W}$, where $MS_B$ is the mean square between the groups and $MS_W$ is the mean square within the groups.

\textbf{Models.} In this study, several machine learning models were applied to predict three-year mortality after percutaneous coronary intervention (PCI). Logistic Regression was used as a baseline method, providing interpretable coefficients that allow estimation of the contribution of individual risk factors, though it is limited by its assumption of linear relationships. To capture more complex dependencies, we employed ensemble-based approaches. Random Forest \cite{ho1995random} constructs multiple decision trees on random subsets of data and features, aggregating their predictions to achieve robust performance and reduced overfitting, while also providing measures of feature importance. Gradient boosting methods were represented by CatBoost and XGBoost. CatBoost is particularly efficient for handling categorical variables and incorporates advanced regularization techniques, often yielding strong performance on structured clinical data. XGBoost is a widely adopted and highly optimized gradient boosting algorithm that offers high predictive accuracy, scalability, and regularization, making it a standard choice in biomedical data analysis.

To ensure the reliability of probabilistic predictions for subsequent analysis, the outputs of tree-based models (including ensemble methods) were further calibrated. This was achieved using the CalibratedClassifierCV implementation from the scikit-learn library, which employs cross-validation to improve the robustness of calibration. Specifically, Platt’s sigmoid scaling was applied, as it has proven effective in correcting systematic biases in probability estimates generated by tree-based classifiers. This approach enhanced the alignment between predicted probabilities and observed event frequencies, which is particularly important in medical prediction tasks where the interpretability and clinical applicability of probability estimates are critical.

\begin{figure}[H]
	\noindent
	\centering
	\includegraphics[width=10.6cm]{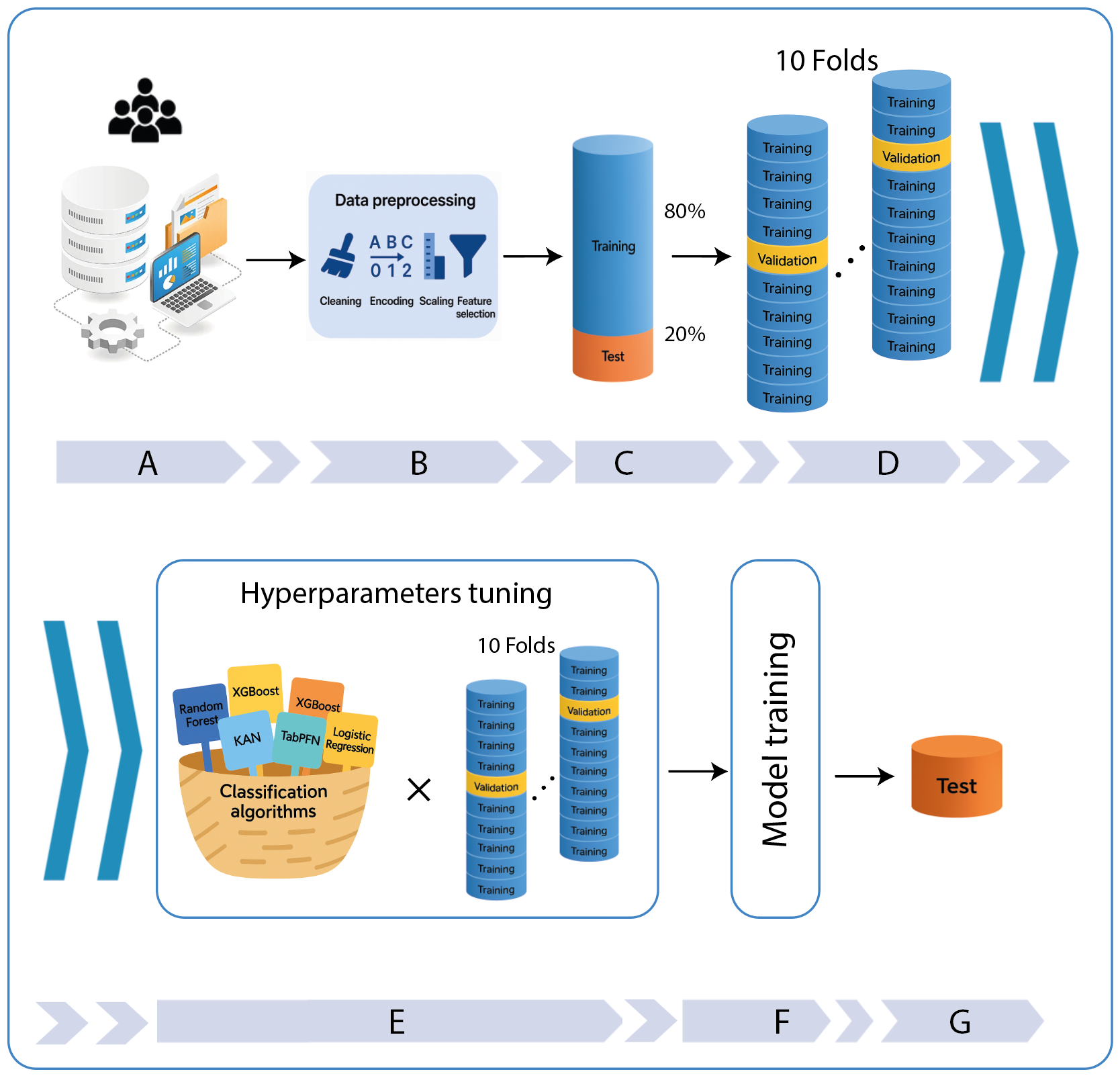}
	\caption{The pipeline of research. a) Data storage and collection  b) Prepocessing - data cleaning, embeddings and etc. c) Train test split  d) 10 folds datasets for cross-validation e) Hyperparameters tuning and selection of the best f) Models training g) Evaluation on test set}
        
	\label{fig:Pipeline}
\end{figure}

In addition to these well-established methods, we also explored two more recent approaches. TabPFN (Tabular Prior-Data Fitted Networks) is a transformer-based model pretrained on thousands of small synthetic datasets to approximate Bayesian inference on tabular data, enabling it to capture complex dependencies with minimal task-specific training. It is worth noting that TabPFN was purpose-built for small-dataset classification and scales poorly as dataset size increases \cite{hollmann2023tabpfn}. Also, this approach excels on datasets with only numerical features and is less strong for datasets with categorical features and missing values. Kolmogorov–Arnold Networks (KAN) \cite{liu2024kan} represent a novel neural network architecture inspired by the Kolmogorov–Arnold representation theorem, allowing efficient approximation of multivariate functions while potentially improving interpretability compared to conventional deep neural networks.

Together, this diverse set of models—from interpretable linear methods to state-of-the-art neural and ensemble approaches—was selected to provide a comprehensive evaluation of machine learning strategies for risk prediction in patients after PCI.

For all models, hyperparameters were optimized during the cross-validation stage using the Tree-structured Parzen Estimator (TPE) algorithm implemented in the Hyperopt library. The selection criterion was the best mean value of the area under the receiver operating characteristic curve (AUC-ROC) across validation folds without synthetic oversampling. 

\textbf{Synthetic data generation}
To address class imbalance and improve the representation of the minority class (patients who died within three years), an additional 500 synthetic samples were generated and added to the training set. This was achieved using several generative algorithms. ARF (Adversarial Random Forest oversampling) \cite{watson2023adversarial}, which dynamically generates synthetic instances by leveraging two ensembles of random forest classifiers - generator forest and discriminator forest to approximate the underlying distribution of the minority class. TVAE (Tabular Variational Autoencoder) \cite{xu2019modeling}, which consists of two neural networks where the encoder learns a probabilistic latent representation of the data and the decoder produces new samples by decoding points drawn from the latent space. TVAE does assume a Gaussian structure in the latent space and Gaussian mixtures for continuous variables. CTGAN (Conditional Tabular GAN) \cite{xu2019modeling}, an adversarial framework specifically designed for tabular data that models complex conditional distributions, non-linear feature interactions, particularly effective for handling imbalanced categorical features. CTGAN is composed of two deep neural networks - generator that creates new synthetic samples and discriminator that tries to differentiate real data from artificial one. GaussianCopula, a statistical method that captures dependencies between variables by modeling their joint distribution through copula functions, enabling realistic sample generation. The model assumes the dependencies follow a normal distribution, and it primarily captures linear dependencies. By combining these approaches, we aimed to enhance the diversity and representativeness of the minority class while reducing the risk of overfitting to synthetic patterns. We also applied TabSyn \cite{zhang2024tabsyn}, novel method for synthetic data generation. It encodes mixed-type data into a continuous latent space using VAE and trains a score-based diffusion model to learn the distribution of those latent embeddings.

During the cross-validation stage, hyperparameters were tuned for the models without the use of synthetic data, and the optimal configurations were selected based on the best average ROC-AUC score. For the Logistic Regression model, the regularization coefficient, the maximum number of training iterations, and the solver were optimized. For the Random Forest model, the tuned parameters included the number of trees in the ensemble, the maximum tree depth, the minimum number of samples required at a leaf node and for a split, the maximum number of features considered for a split, and the use of bootstrap sampling. For the CatBoost model, the optimized hyperparameters included the number of trees in the ensemble, tree depth, learning rate, the L2 regularization coefficient, and the bootstrap type. For the XGBoost model, the following hyperparameters were subject to optimization: the number of estimators, maximum tree depth, learning rate, minimum child weight, subsample ratio, and column subsample ratio by tree. For the KAN model, the hyperparameters subject to tuning included the grid size, spline order, the learning rate, and the weight assigned to samples in the Binary Cross-Entropy loss function.

Preprocessing and feature scaling (see figure \ref{fig:Pipeline}) procedures were applied exclusively to the training dataset in order to prevent data leakage into the validation and test sets. This ensured that all transformations, including normalization and scaling, were fitted solely on the training data and subsequently applied to the validation and test data, thereby preserving the integrity and fairness of the model evaluation.

To evaluate model performance, three training and assessment strategies were employed in order to obtain confidence intervals for the performance metrics. First, the model was trained on clean data without synthetic samples and evaluated using 10-fold cross-validation. Second, 500 synthetic samples from the positive class were added to the training folds, and the model’s performance was assessed on the corresponding test folds. Third, edge cases were incorporated into the training data within the 10-fold cross-validation setting, and the models were subsequently evaluated on the test folds. These scenarios allowed us to estimate the variability of performance and to obtain metric intervals reflecting model robustness.

\begin{figure}[H]
	\noindent
	\centering
	\includegraphics[width=12.6cm]{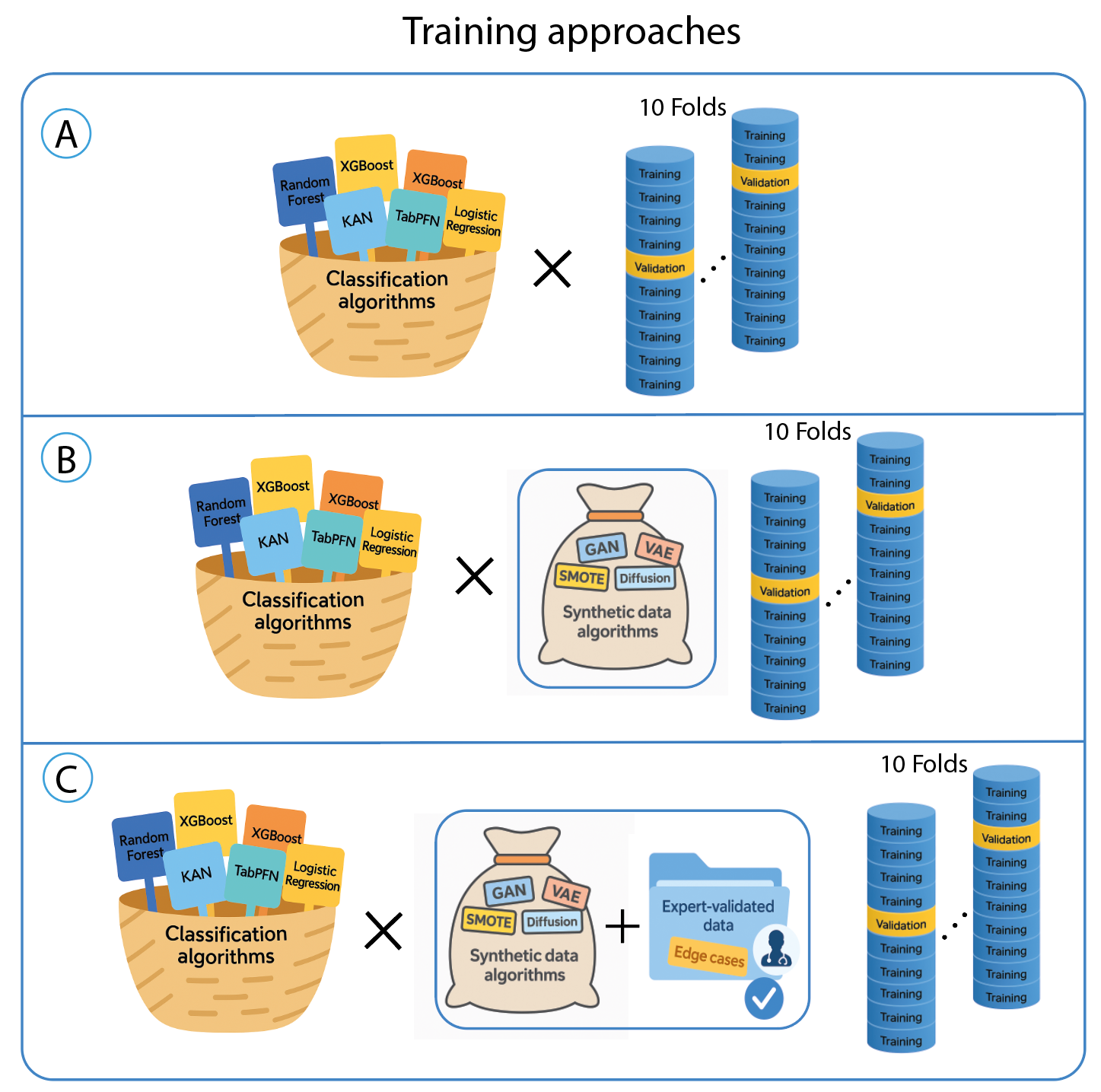}
	\caption{Three training regimes: a) Standard 10 folds cross validation. b) Synthetic in distribution data used to enrich the dataset. c) Synthetic data and expert validated edge cases.}
        
	\label{fig:training_regimes}
\end{figure}

\textbf{Edge-case design.} Since machine learning models are known to perform poorly when extrapolating to data points that fall outside the distribution of the training set, we additionally evaluated model performance on generated edge cases. These edge cases were sampled to reflect clinical scenarios in which the patient’s condition is critical, representing extreme values of key clinical features (Table~\ref{table:features_dataset}). 

This approach allows us to assess the robustness and reliability of the models under clinically challenging conditions that are underrepresented in the original dataset. Edge cases refer to a dataset specifically designed to evaluate the robustness of models in predicting three-year mortality among critically ill patients. For this task, the features used in the original dataset were constrained to take clinically critical values with high probability (0.8–0.9 for categorical features). The data were generated manually from predefined distributions. To facilitate testing on edge cases, a cohort of 200 synthetic patients was created, and all trained models were evaluated on this set.

In addition to their role in stress-testing the models, the "edge cases" as it schematically presented in figure \ref{fig:training_regimes}, were also incorporated into the training process to partially mitigate class imbalance. By enriching the dataset with critical patient profiles, we aimed both to improve minority class representation and to ensure that the models remain reliable when applied to high-risk clinical subgroups. 

\textbf{Feature importance scoring} 
To evaluate the contribution of individual features to model performance, we applied permutation feature importance, chosen for its universality across all presented models. In this procedure, the values of a given feature were randomly permuted across samples,
and the resulting decrease in predictive performance was quantified using the AUC-ROC metric. Each feature permutation was repeated 10 times, and the mean decrease in AUC-ROC was taken as a stable estimate of the feature’s importance.

\textbf{External validation} 
For external validation of the models, a dataset of 158 patients (positive class = 29) from the Tyumen Cardiology Research Center, Tomsk National Research Medical Center, Russian Academy of Sciences who underwent PCI for acute myocardial infarction for the period 2011 to 2013.
\section{Results}

\textbf{Baseline}. Without oversampling, all models achieve high overall accuracy (0.92-0.93), yet they almost completely ignore the minority class (see Table~\ref{table:test_class_metr}). Logistic Regression, Random Forest, CatBoost, and TabPFN obtain zero values for F1-score, Precision, and Recall, while their AUC-ROC values remain relatively high (0.76–0.82). This indicates that although the models are able to rank instances to some extent, they fail to identify the minority class at the decision threshold (0.5). Only KAN and XGBoost achieve non-zero F1-scores (\~0.06), with perfect precision but extremely low recall. In this setting, the calibration metrics appear deceptively good (Table 2): the average predicted confidence is approximately 0.92 with Expected Calibration Error (ECE) below 0.04. This, however, is a reflection of the models’ tendency to consistently predict the majority class with high confidence rather than genuine calibration.

\begin{table}[H]
    \caption{Classification metrics on test set and probability distribution on edge cases}
    \label{validation}
    \centering
    \setlength{\tabcolsep}{4pt}
    \begin{tabular}{llllllllllll}
    \toprule
        {Model} & {AUC-ROC} & {F1-score} & {Precision} & {Recall} & {Accuracy} & {Q0} & {Q50} & {Q99} & {Mean} & {Std} \\
        \midrule
        \multicolumn{6}{c}{Models without oversampling} \\
        \hline
        \ce{Logistic Regression} & 0.78 & 0 & 0 &  0 & 0.92 &  0.16 & 0.23 & 0.30 & 0.23 & 0.03\\
        \hline
        \ce{Random Forest} & 0.78 & 0 & 0 &  0 & 0.92 & 0.34 & 0.51 & 0.58 & 0.51 & 0.04\\
        \hline
        \ce{CatBoost} & 0.76 & 0 & 0 & 0 & 0.92 &  0.33 & 0.63 & 0.74 & 0.61 & 0.07\\
        \hline
        \ce{TabPFN} & \textbf{0.82} & 0 & 0 & 0 & 0.92 & 0.46 & 0.65 & 0.72 & 0.64 & 0.04\\
        \hline
        \ce{KAN} & 0.79 & \textbf{0.06} & \textbf{1} & \textbf{0.03} & \textbf{0.93} & 0.68 & 0.99 & 0.99 & 0.98 & 0.03\\
        \hline
        \ce{XGBoost} & 0.79 & \textbf{0.06} & \textbf{1} & \textbf{0.03} & \textbf{0.93}  & 0.40 & 0.64 & 0.74 & 0.62 & 0.06\\
        \hline
        \multicolumn{6}{c}{Models with ARF oversampling} \\
        \hline
        \ce{Logistic Regression} & 0.77 & 0.06 & 0.5 &  0.03 & \textbf{0.92}  & 0.53 & 0.71 & 0.81 & 0.71 & 0.05 \\
        \hline
        \ce{Random Forest} & \textbf{0.78} & \textbf{0.29} & 0.28 & 0.29 & 0.89 & 0.74 & 0.88 & 0.93 & 0.88 & 0.03\\
        \hline
        \ce{CatBoost} & 0.77 & 0.14 & 0.23 & 0.1 & 0.90 & 0.83 & 0.96 & 0.97 & 0.95 & 0.02\\
        \hline
        \ce{TabPFN} & 0.71 & 0.05 & 0.17 & 0.03 & 0.91 & 0.75 & 0.97 & 0.99 & 0.95 & 0.04\\
        \hline
        \ce{KAN} & 0.75 & 0.27 & 0.18 & \textbf{0.55} & 0.78 & 0.56 & 0.99 & 0.99 & 0.96 & 0.05\\
        \hline
        \ce{XGBoost} & 0.75 & 0.19 & \textbf{0.33} & 0.13 & 0.91 & 0.66 & 0.96 & 0.98 & 0.94 & 0.04\\
        \hline
        
        \multicolumn{6}{c}{Models with Edge cases oversampling} \\
        \hline
        \ce{Logistic Regression} & 0.78 & 0 & 0 & 0 & 0.92 & 0.60 & 0.97 & 0.99 & 0.95 & 0.05\\
        \hline
        \ce{Random Forest} & 0.78 & 0 & 0 & 0 & 0.92 &  0.98 & 0.99 & 0.99 & 0.99 & 0.00\\
        \hline
        \ce{CatBoost} & 0.77 & 0.12 & 0.67 & 0.06 & \textbf{0.93} & 0.98 & 0.98 & 0.98 & 0.98 & 0.00\\
        \hline
        \ce{TabPFN} & \textbf{0.81} & 0.06 & \textbf{1} & 0.03 & \textbf{0.93} & 0.99 & 0.99 & 0.99 & 0.99 & 0.00\\
        \hline
        \ce{KAN} & 0.76 & \textbf{0.14} & 0.27 & \textbf{0.1} & 0.91 & 0.92 & 0.99 & 0.99 & 0.99 & 0.01\\
        \hline
        \ce{XGBoost} & 0.79 & 0.06 & 0.5 & 0.03 & 0.92 & 0.99 & 0.99 & 0.99 & 0.99 & 0.00\\
        
    \bottomrule
    \end{tabular}
    \label{table:test_class_metr}

\end{table}

\textbf{Effect of oversampling methods}.
The behavior of models changes when synthetic data are added to the training set. Some of the most notable improvements are observed for models trained with ARF-based synthetic data (Table ~\ref{table:test_class_metr}). On the test set, these models achieve relatively strong performance, with AUC-ROC values ranging from 0.75 to 0.78. Although these results are slightly lower than those obtained when training on the original dataset without oversampling, the models demonstrate clear gains on other metrics. For example, Random Forest maintains an AUC-ROC score of 0.78 while improving its F1-score to 0.29, with Recall also increasing to 0.29. The KAN model, when augmented with ARF-generated data, reaches a Recall of 0.55. Overall, these results indicate that the models become more effective at detecting instances of the minority class. The benefits of ARF synthetic data are further confirmed in cross-validation experiments (Table ~\ref{table:cv_class_metr}), where models achieve mean AUC-ROC values of 0.67–0.72. While this reflects a slight decrease in the primary metric, substantial improvements are observed in F1, Precision, and Recall.

Models trained with GAN-generated synthetic data also demonstrate strong performance on the test set, achieving AUC-ROC values of 0.75–0.78 (Table ~\ref{table:test_class_metr}). The KAN model achieves an F1-score of 0.30 and an AUC-ROC of 0.76, while Random Forest and CatBoost ensembles both reach F1-scores of approximately 0.29. However, it is important to note that the mean AUC-ROC across models trained with GAN-generated data is 0.63–0.67, which is lower than the results obtained with ARF-based augmentation.

TVAE-based synthetic data yield more moderate results. On the test set, models trained with TVAE augmentation achieve AUC-ROC values between 0.58 and 0.73 (Table ~\ref{table:test_class_metr}). Both Random Forest and KAN record F1-scores of 0.16, which is over 10\% lower than those observed with ARF and GAN. Cross-validation results (Table ~\ref{table:cv_class_metr}) further indicate a decline in average metrics compared to ARF-based experiments.

Experiments with Gaussian Copula synthetic data show that Random Forest achieves an AUC-ROC of 0.78, a relatively strong score, although its F1-score remains moderate at 0.20. The KAN model demonstrates an F1-score of 0.21 and the highest Recall among all models at 0.58, though its AUC-ROC decreases to 0.73. Cross-validation further highlights the average nature of these results, with Random Forest reaching a mean AUC-ROC of 0.70 and Precision of 0.21, while KAN achieves Recall of 0.44.

We also experimented with TabSyn, a diffusion-based generative model for tabular data. This method did not yield competitive results for this highly imbalanced classification task with a limited dataset. Most models showed only marginal improvements in F1-score (0.07–0.09), with the exception of KAN, which achieved 0.21. However, this value still falls short of results obtained with ARF, Gaussian Copula, and GAN. Furthermore, AUC-ROC values (0.75–0.76) were consistently lower than those achieved with ARF and GAN.

\textbf{Edge-case stress test}
Additionally, we conducted experiments using synthetic edge cases. The rationale was that such data could help models identify critical cases, specifically patients with a confirmed fatal outcome within three years. When added to the training set, these synthetic samples improved KAN’s metrics to F1 = 0.14, Recall = 0.10, and Precision = 0.14. Although the improvements are modest compared to other approaches, they enhance model performance without requiring additional generative training. On the test set, AUC-ROC values ranged from 0.76 to 0.81. In cross-validation, KAN achieved an average F1-score of 0.18 and the highest Precision across all models at 0.32.

\begin{table}[H]
    \caption{Classification metrics on Cross validation}
    \label{validation}
    \centering
    \setlength{\tabcolsep}{10pt}
    \begin{tabular}{llllll}
    \toprule
        {Model} & {AUC-ROC} & {F1-score} & {Precision} & {Recall} & {Accuracy} \\
        \midrule
        \multicolumn{6}{c}{Models without oversampling} \\
        \hline
        \ce{Logistic Regression} & $\textbf{0.72} \pm 0.11$ & $0.00 \pm 0.00$ & $0.00 \pm 0.00$ & $0.00 \pm 0.00$ & $0.92 \pm 0.00$ \\
        \hline
        \ce{Random Forest} & $\textbf{0.72} \pm 0.11$ & $0.00 \pm 0.00$ &  $0.00 \pm 0.00$ &  $0.00 \pm 0.00$ & $0.92 \pm 0.00$ \\
        \hline
        \ce{CatBoost} & $0.70 \pm 0.11$ & $\textbf{0.03} \pm 0.06$ & $0.15 \pm 0.34$ & $\textbf{0.02} \pm 0.03$ & $0.92 \pm 0.00$ \\
        \hline
        \ce{TabPFN} & $0.70 \pm 0.12$ & $0.00 \pm 0.00$ & $0.00 \pm 0.00$ & $0.00 \pm 0.00$ & $0.92 \pm 0.00$ \\
        \hline
        \ce{KAN} & $\textbf{0.72} \pm 0.12$ & $\textbf{0.03} \pm 0.06$ & $\textbf{0.2} \pm 0.42$ & $\textbf{0.02} \pm 0.03$ & $0.92 \pm 0.01$ \\
        \hline
        \ce{XGBoost} & $\textbf{0.72} \pm 0.12$ & $\textbf{0.03} \pm 0.06$ & $\textbf{0.2} \pm 0.42$ & $\textbf{0.02} \pm 0.03$ & $0.92 \pm 0.00$ \\
        \hline
        \multicolumn{6}{c}{Models with ARF oversampling} \\
        \hline
        \ce{Logistic Regression} & $\textbf{0.72} \pm 0.11$ & $0.09 \pm 0.12$ & $\textbf{0.26} \pm 0.35$ & $0.06 \pm 0.07$ & $\textbf{0.92} \pm 0.01$ \\
        \hline
        \ce{Random Forest} & $0.71 \pm 0.10$ & $0.22 \pm 0.15$ & $0.18 \pm 0.12$ &  $0.28 \pm 0.22$ & $0.86 \pm 0.02$ \\
        \hline
        \ce{CatBoost} & $0.71 \pm 0.11$ & $0.17 \pm 0.14$ & $0.21 \pm 0.15$ & $0.16 \pm 0.15$ & $0.9 \pm 0.02$ \\
        \hline
        \ce{TabPFN} & $0.67 \pm 0.11$ & $0.06 \pm 0.06$ & $0.13 \pm 0.16$ & $0.04 \pm 0.04$ & $0.9 \pm 0.02$ \\
        \hline
        \ce{KAN} & $\textbf{0.72} \pm 0.11$ & $\textbf{0.24} \pm 0.09$ & $0.15 \pm 0.06$ & $\textbf{0.54} \pm 0.18$ & $0.73 \pm 0.05$\\
        \hline
        \ce{XGBoost} & $0.69 \pm 0.10$ & $0.18 \pm 0.16$ & $0.22 \pm 0.20$ & $0.17 \pm 0.16$ & $0.89 \pm 0.02$ \\
        \hline
        
        \multicolumn{6}{c}{Models with Edge cases oversampling} \\
        \hline
        \ce{Logistic Regression} & $0.7 \pm 0.11$ & $0.03 \pm 0.08$ & $0.07 \pm 0.2$ & $0.02 \pm 0.05$ & $\textbf{0.92} \pm 0.01$ \\
        \hline
        \ce{Random Forest} & $0.71 \pm 0.11$ & $0.01 \pm 0.04$ & $0.05 \pm 0.15$ &  $0.01 \pm 0.02$ & $\textbf{0.92} \pm 0$ \\
        \hline
        \ce{CatBoost} & $0.71 \pm 0.1$ & $0.05 \pm 0.07$ & $0.23 \pm 0.33$ & $0.03 \pm 0.04$ & $\textbf{0.92} \pm 0$ \\
        \hline
        \ce{TabPFN} & $0.70 \pm 0.1$ & $0 \pm 0$ & $0 \pm 0$ & $0 \pm 0$ & $\textbf{0.92} \pm 0$ \\
        \hline
        \ce{KAN} & $0.70 \pm 0.09$ & $\textbf{0.18} \pm 0.15$ & $\textbf{0.32} \pm 0.32$ & $\textbf{0.13} \pm 0.11$ & $0.91 \pm 0.02$\\
        \hline
        \ce{XGBoost} & $\textbf{0.72} \pm 0.11$ & $0.03 \pm 0.06$ & $0.15 \pm 0.32$ & $0.02 \pm 0.03$ & $\textbf{0.92} \pm 0$ \\
    \bottomrule
    \end{tabular}
    \label{table:cv_class_metr}
\end{table}

\textbf{Calibration}
Calibration metrics highlight another important trade-off Table~\ref{table:test_prob_metr} and Table~\ref{table:cv_prob_metr}. Logistic Regression becomes poorly calibrated after ARF or TVAE oversampling (ECE > 0.2, average confidence $\approx$ 0.7). Ensemble methods (Random Forest, CatBoost, XGBoost) retain more stable calibration (Confidence $\approx$ 0.85–0.89, ECE $\approx$ 0.05–0.1). TabPFN remains the most calibrated model across all settings (ECE < 0.03), yet its Recall remains consistently close to zero, showing that well-calibrated models are not necessarily effective in detecting minority-class instances. Another noteworthy observation is that the inclusion of synthetic data tends to reduce the average confidence of the models. Without synthetic augmentation, the mean confidence remains around 0.92; however, after incorporating synthetic data, this value decreases.

\begin{table}[H]
    \caption{Probabilistic metrics on test set}
    \label{validation}
    \centering
    \setlength{\tabcolsep}{6pt}
    \begin{tabular}{llllll}
    \toprule
        {Model} & {Avg Confidence} & {Avg Entropy} & {Brier score} & {ECE}\\
        \midrule
        \multicolumn{4}{c}{Models without oversampling} \\
        \hline
        \ce{Logistic Regression} & 0.92 & 0.27 & 0.07 & 0.03 \\
        \hline
        \ce{Random Forest} & 0.92 & 0.26 & 0.06 & 0.03\\
        \hline
        \ce{CatBoost} & 0.92 & 0.27 & 0.06 & 0.02\\
        \hline
        \ce{TabPFN} & 0.93 & 0.22 & 0.06 & 0.03\\
        \hline
        \ce{KAN} & 0.92 & 0.26 & 0.06 & 0.04\\
        \hline
        \ce{XGBoost} & 0.92 & 0.26 & 0.06 & 0.04\\
        \hline
        \multicolumn{4}{c}{Models with ARF oversampling} \\
        \hline
        \ce{Logistic Regression} & 0.72 & 0.59 & 0.11 & 0.21 \\
        \hline
        \ce{Random Forest} & 0.83 & 0.40 & 0.09 & 0.11\\
        \hline
        \ce{CatBoost} & 0.88 & 0.33 & 0.07 & 0.06\\
        \hline
        \ce{TabPFN} & 0.93 & 0.24 & 0.07 & 0.02\\
        \hline
        \ce{KAN} & 0.69 & 0.6 & 0.18 & 0.31\\ 
        \hline
        \ce{XGBoost} & 0.88 & 0.34 & 0.07 & 0.06\\
        \hline
        
        \multicolumn{4}{c}{Models with Edge cases oversampling} \\
        \hline
        \ce{Logistic Regression} & 0.91 & 0.29 & 0.06 & 0.04 \\
        \hline
        \ce{Random Forest} &  0.92 & 0.27 & 0.06 & 0.04\\
        \hline
        \ce{CatBoost} & 0.92 & 0.26 & 0.06 & 0.03 \\
        \hline
        \ce{TabPFN} & 0.94 & 0.2 & 0.06 & 0.03\\
        \hline
        \ce{KAN} & 0.88 & 0.32 & 0.07 & 0.05\\ 
        \hline
        \ce{XGBoost} & 0.92 & 0.27 & 0.06 & 0.05\\
    \bottomrule
    \end{tabular}
    \label{table:test_prob_metr}

\end{table}

\begin{table}[H]
    \caption{Probabilistic metrics on Cross validation}
    \label{validation}
    \centering
    \setlength{\tabcolsep}{10pt}
    \begin{tabular}{llllll}
    \toprule
        {Model} & {Avg Confidence} & {Avg Entropy} & {Brier score} & {ECE}\\
        \midrule
        \multicolumn{4}{c}{Models without oversampling} \\
        \hline
        \ce{Logistic Regression} & $0.92 \pm 0$ & $0.27 \pm 0$ & $0.07 \pm 0$ & $0.03 \pm 0.03$ \\
        \hline
        \ce{Random Forest} & $0.92 \pm 0.01$ & $0.26 \pm 0.01$ & $0.07 \pm 0.01$ & $0.03 \pm 0.02$ \\
        \hline
        \ce{CatBoost} & $0.92 \pm 0.01$ & $0.26 \pm 0.01$ & $0.07 \pm 0.01$ & $0.02 \pm 0.02$ \\
        \hline
        \ce{TabPFN} & $0.93 \pm 0.01$ & $0.22 \pm 0.01$ & $0.07 \pm 0.01$ & $0.03 \pm 0.01$\\
        \hline
        \ce{KAN} & $0.92 \pm 0.01$ & $0.26 \pm 0.01$ & $0.07 \pm 0.01$ & $0.03 \pm 0.02$ \\
        \hline
        \ce{XGBoost} & $0.92 \pm 0.01$ & $0.26 \pm 0.01$ & $0.07 \pm 0.01$ & $0.03 \pm 0.02$ \\
        \hline
        \multicolumn{4}{c}{Models with ARF oversampling} \\
        \hline
        \ce{Logistic Regression} & $0.70 \pm 0.01$ & $0.6 \pm 0.01$ & $0.12 \pm 0.01$ & $0.23 \pm 0.01$ \\
        \hline
        \ce{Random Forest} & $0.82 \pm 0.01$ & $0.42 \pm 0.02$ & $0.10 \pm 0.01$ & $0.14 \pm 0.02$ \\
        \hline
        \ce{CatBoost} & $0.88 \pm 0.01$ & $0.32 \pm 0.01$ & $0.09 \pm 0.02$ & $0.07 \pm 0.01$ \\
        \hline
        \ce{TabPFN} & $0.92 \pm 0$ & $0.24 \pm 0.01$ & $0.08 \pm 0.01$ & $0.06 \pm 0.02$\\
        \hline
        \ce{KAN} & $0.67 \pm 0.01$ & $0.60 \pm 0.01$ & $0.19 \pm 0.02$ & $0.33 \pm 0.02$ \\
        \hline
        \ce{XGBoost} & $0.87 \pm 0.01$ & $0.33 \pm 0.01$ & $0.09 \pm 0.02$ & $0.09 \pm 0.02$ \\
        \hline
        
        \multicolumn{4}{c}{Models with Edge cases oversampling} \\
        \hline
        \ce{Logistic Regression} & $0.91 \pm 0$ & $0.3 \pm 0.01$ & $0.07 \pm 0.01$ & $0.04 \pm 0.02$ \\
        \hline
        \ce{Random Forest} & $0.92 \pm 0$ & $0.26 \pm 0.01$ & $0.07 \pm 0.01$ & $0.03 \pm 0.02$ \\
        \hline
        \ce{CatBoost} & $0.92 \pm 0.01$ & $0.26 \pm 0.01$ & $0.07 \pm 0.01$ & $0.03 \pm 0.02$ \\
        \hline
        \ce{TabPFN} & $0.94 \pm 0$ & $0.2 \pm 0.01$ & $0.07 \pm 0.01$ & $0.04 \pm 0.01$ \\
        \hline
        \ce{KAN} & $0.88 \pm 0.01$ & $0.31 \pm 0.01$ & $0.08 \pm 0.01$ & $0.06 \pm 0.01$\\
        \hline
        \ce{XGBoost} & $0.92 \pm 0$ & $0.27 \pm 0.01$ & $0.07 \pm 0$ & $0.03 \pm 0.02$ \\
        
    \bottomrule
    \end{tabular}
    \label{table:cv_prob_metr}
\end{table}

\textbf{Evaluation}
One of the evaluation stages involved testing the models on critical patients (edge cases). Our goal was to ensure that the models assign a higher probability of mortality to these patients. If a model exhibits high variability or systematically underestimates mortality in this setting, then—regardless of its test-set performance metrics—it cannot be considered suitable for practical applications.

\begin{figure}[H]
	\noindent
	\centering
	\includegraphics[width=10.6cm]{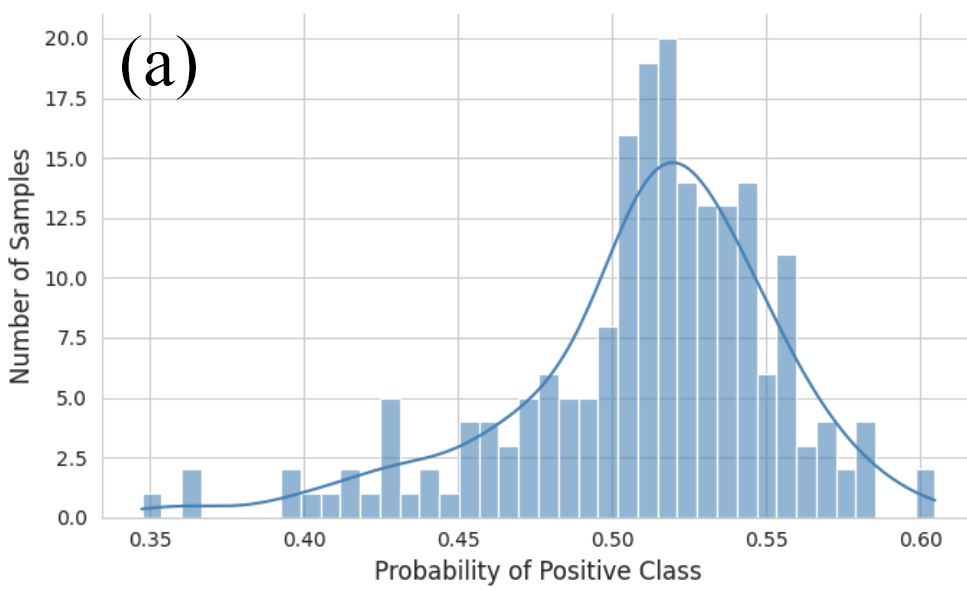}
        \includegraphics[width=10.6cm]{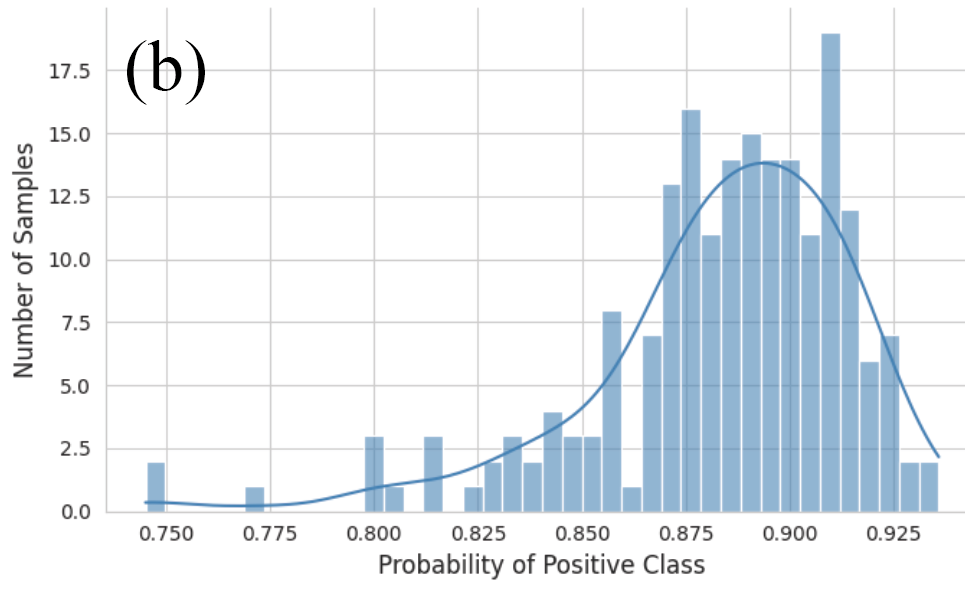}
	\caption{Random Forest distribution probabilities on edge cases without oversampling (a) and with ARF (b).}
        
	\label{fig:Random_Forest_distribution}
\end{figure}

The most informative results emerge from the evaluation on edge cases (see Table~\ref {table:test_class_metr_full}), where all samples belong to the positive class. Ideally, models should assign high probabilities in this scenario. Without oversampling, Logistic Regression (mean 0.23) and Random Forest assign low probabilities (mean 0.51 see figure ~\ref{fig:Random_Forest_distribution} (a)), indicating a failure to recognize critical cases. CatBoost, XGBoost, and TabPFN perform moderately better (mean 0.51–0.64), while KAN is the only model to output near-optimal predictions with a mean probability of 0.98 and a standard deviation of 0.03. After applying oversampling, performance improves across models: CatBoost (mean 0.95) and Random Forest with ARF reach mean probabilities of 0.88 (see figure~\ref{fig:Random_Forest_distribution} (b)), respectively, and TabPFN with GAN reaches 0.98. Nevertheless, KAN consistently produces the highest probabilities for edge cases (0.94–0.96).

\begin{figure}[H]
    \centering
    \begin{subfigure}{0.48\textwidth}
        \centering
        \includegraphics[width=7.5cm]{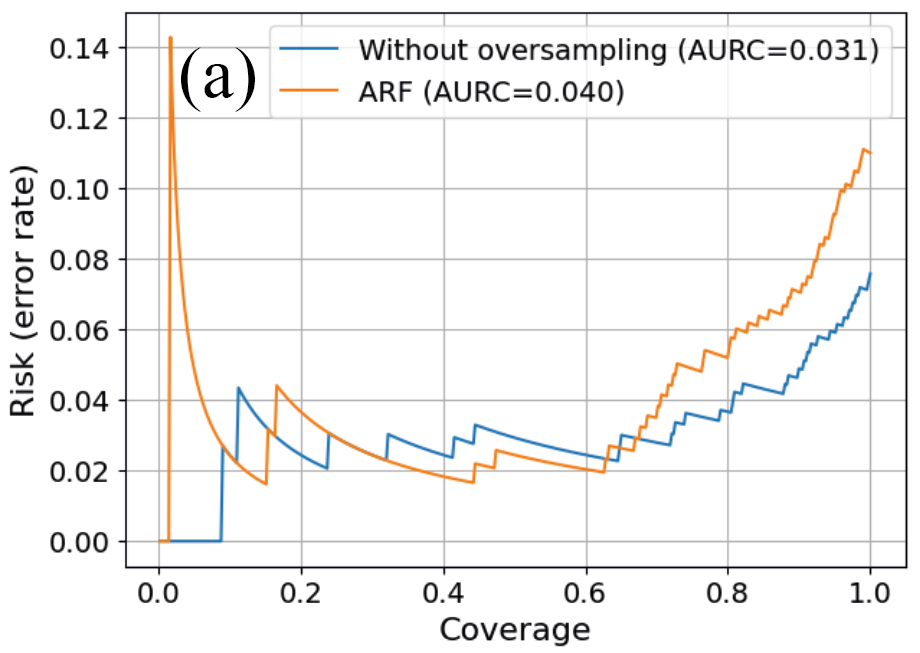}
        
        \label{fig:rc_rf}
    \end{subfigure}
    \hfill
    \begin{subfigure}{0.48\textwidth}
        \includegraphics[width=7.5cm]{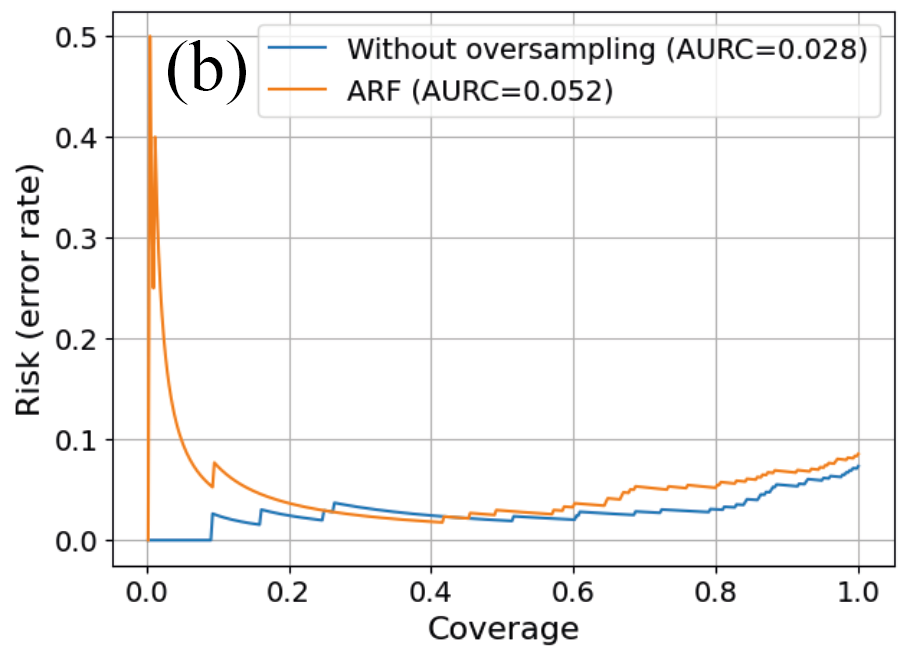}
        
        \label{fig:rc_xgb}
    \end{subfigure}
    \caption{Risk coverage analysis on the test set across different models. (a) Random Forest (b) XGBoost}
    \label{fig:risk_coverage}
\end{figure}

\textbf{Risk–Coverage.} Risk–Coverage (RC) curves were constructed for two models: Random Forest and XGBoost (see figure \ref{fig:risk_coverage}). The curves show the performance of each model on the test set, both with the baseline data and after applying the ARF oversampling method.

Both models achieve an AUC-RC (Risk–Coverage) value below 0.1, which indicates that they are able to perform effective selective prediction. This result demonstrates that both the oversampled and the non-oversampled versions of the models can reliably assess prediction confidence and maintain a low risk of errors when making selective predictions.

\textbf{Feature Importance.} According to feature importance analysis (Table ~\ref{table:features_importance}), four features emerged as the most influential: Age, Ejection Fraction, Peripheral Artery Disease, and Cerebrovascular Disease. The models frequently relied on these variables when making predictions on the test set. A notable observation was that the models almost entirely disregarded the CKD feature, despite its clinical relevance as a key indicator of kidney function.
Building on this, we further present the averaged feature importance rankings across all models for each type of synthetic data. These rankings are expressed as indices reflecting the relative importance of each feature, with rank 1 indicating the most important and the highest index corresponding to the least important. Averaging across models provides a more robust assessment of feature relevance, ensuring a consistent comparison across synthetic data generation methods.

\begin{table}[H]
    \caption{Feature importance}
    \centering
    \setlength{\tabcolsep}{10pt}
    \begin{tabular}{llllll}
    \toprule

        {Feature} & {Without oversampling} & {ARF}\\
        \midrule
        \ce{Age} & 4.66 & 2 \\
        \hline
        \ce{Anemia} & 8.33 & 6.83\\
        \hline
        \ce{Ejection Fraction} &  4.16 & 1 \\
        \hline
        \ce{Cerebrovascular Disease} & 4.5 & 4 \\
        \hline
        \ce{CKD} & 17.5 & 18.0\\
        \hline
        \ce{Peripheral Artery Disease } & 5.0 & 3.33\\
        \hline
        \ce{Aortic Stenosis} & 11.0 & 10.0\\
        \hline
        \ce{Single Vessel disease} & 4.17 & 5.33\\
        \hline
        \ce{Coronary calcium} & 8.0 & 11.66\\
        \hline
        \ce{Stent type - Calypso} & 11.83 & 11.33\\
        \hline
        \ce{Medina side} & 19.8 & 17.0\\
        \hline
        \ce{Atrial Fibrillation} & 13.66 &  12.83\\
        \hline
        \ce{DEFINITION score} & 9.67 & 11.67\\
        \hline
        \ce{History of cancer} & 15.0 & 14.5\\
        \hline
        \ce{Stent type - Synergy} & 13.0 & 10.5\\
        \hline
        \ce{Stent diameter} & 14.83 & 18.0\\
        \hline
        \ce{Stent length} & 14.5 & 11.17\\
        \hline
        \ce{Ad-hoc PCI} & 6.16 & 7.83 \\
        \hline
        \ce{Previous PCI} & 16.33 & 18.17\\
        \hline
        \ce{Stent type - Xience} & 12.5 & 15.83\\
        \hline
        \ce{CTO bifurcation} & 16.33 & 20.0\\
    \bottomrule
    \end{tabular}
    \label{table:features_importance}
\end{table}

\textbf{Training based on feature importance.} Another important finding from the experiments was that some features turned out to be entirely irrelevant and had no impact on the model’s ability to predict long-term mortality. Therefore, an additional experiment was conducted to evaluate how the model’s predictions would change after removing these non-informative features from the training and test datasets.

As a result, a dataset was constructed that included the following features: Age, Anemia, Ejection fraction, Cerebrovascular disease, CKD, Peripheral artery disease, Aortic stenosis, Single vessel disease, Coronary calcium, Atrial fibrillation, History of cancer, Ad-hoc PCI, Previous PCI. The criterion for feature selection in this experiment was the permutation importance measure. In addition, all clinical features from the original dataset were retained.

The training was carried out using synthetic data generation with ARF, and the resulting models were evaluated both on the test dataset and on an external validation dataset. When using these features, all models demonstrated improved performance on the dataset (Tables ~\ref{table:ext_test_class_metr_feature_imp} and ~\ref{table:ext_test_prob_metr_feature_imp}). The AUC-ROC on the test set exceeded 0.8 for all models. TabPFN achieved an F1 score of 0.26, although none of the synthetic data approaches enabled this model to reach satisfactory performance. Such behavior may indicate a high level of noise in the data of the first experiment and suggest that dimensionality reduction helps focus on the most relevant factors for prediction.

\begin{table}[H]
    \caption{Classification metrics on test and external validation dataset and cross-validation with training based on feature importance}
    \label{validation}
    \centering
    \setlength{\tabcolsep}{10pt}
    \begin{tabular}{llllllllllll}
    \toprule
        {Model} & {AUC-ROC} & {F1-score} & {Precision} & {Recall} & {Accuracy}\\
        \midrule
        \multicolumn{6}{c}{Models on test set} \\
        \hline
        \ce{Logistic Regression} & 0.80 & 0.06 & 0.33 & 0.03 & 0.92\\
        \hline
        \ce{Random Forest} & 0.84 & 0.44 & 0.37 & 0.55 & 0.89\\
        \hline
        \ce{CatBoost} & 0.82 & 0.44 & 0.36 & 0.58 & 0.89\\
        \hline
        \ce{TabPFN} & 0.83 & 0.26 & 0.40 & 0.19 & 0.92\\
        \hline
        \ce{KAN} & 0.82 & 0.34 & 0.22 & 0.74 & 0.78\\
        \hline
        \ce{XGBoost} & 0.84 & 0.43 & 0.36 & 0.52 & 0.89\\
        \hline
        \multicolumn{6}{c}{Models on cross-validation} \\
        \hline
        \ce{Logistic Regression} & $0.71 \pm 0.11$ & $0.14 \pm 0.13$ & $0.38 \pm 0.39$ & $0.09 \pm 0.08$ & $0.92 \pm 0.01$ \\
        \hline
        \ce{Random Forest} & $0.70 \pm 0.11$ & $0.25 \pm 0.13$ & $0.22 \pm 0.10$ & $0.30 \pm 0.17$ & $0.86 \pm 0.03$ \\
        \hline
        \ce{CatBoost} & $0.71 \pm 0.11$ & $0.24 \pm 0.15$ & $0.22 \pm 0.13$ & $0.28 \pm 0.17$ & $0.86 \pm 0.04$\\
        \hline
        \ce{TabPFN} & $0.70 \pm 0.11$ & $0.17 \pm 0.12$ & $0.25 \pm 0.20$ & $0.13 \pm 0.10$ & $0.90 \pm 0.02$\\
        \hline
        \ce{KAN} & $0.71 \pm 0.11$ & $0.24 \pm 0.08$ & $0.16 \pm 0.05$ & $0.55 \pm 0.17$ & $0.73 \pm 0.05$\\
        \hline
        \ce{XGBoost} & $0.71 \pm 0.12$ & $0.26 \pm 0.13$ & $0.23 \pm 0.12$ & $0.31 \pm 0.18$ & $0.86 \pm 0.03$\\
        \hline
        \multicolumn{6}{c}{Models on external validation} \\
        \hline
        \ce{Logistic Regression}  & 0.69 & 0.13 & 1.00 & 0.07 & 0.83 \\
        \hline
        \ce{Random Forest} & 0.71 & 0.42 & 0.46 & 0.38 & 0.80 \\
        \hline
        \ce{CatBoost} & 0.71 & 0.36 & 0.38 & 0.34 & 0.78 \\
        \hline
        \ce{TabPFN} & 0.74 & 0.34 & 0.58 & 0.24 & 0.83\\
        \hline
        \ce{KAN} & 0.69 & 0.42 & 0.33 & 0.59 & 0.71\\
        \hline
        \ce{XGBoost}  & 0.70 & 0.39 & 0.39 & 0.38 & 0.78 \\
    \bottomrule
    \end{tabular}
    \label{table:ext_test_class_metr_feature_imp}

\end{table}

\begin{table}[H]
    \caption{Probability metrics on test and external validation dataset and cross-validation with training based on feature importance}
    \label{validation}
    \centering
    \setlength{\tabcolsep}{10pt}
    \begin{tabular}{lllll}
    \toprule
        {Model} & {Avg Confidence} & {Avg Entropy} & {Brier Score} & {ECE}\\
        \midrule
        \multicolumn{4}{c}{Models on test set} \\
        \hline
        \ce{Logistic Regression} & 0.72 & 0.58 & 0.11 & 0.21 \\
        \hline
        \ce{Random Forest} & 0.77 & 0.50 & 0.10 & 0.18\\
        \hline
        \ce{CatBoost} & 0.76 & 0.51 & 0.10 & 0.19\\
        \hline
        \ce{TabPFN} & 0.83 & 0.41 & 0.07 & 0.10\\
        \hline
        \ce{KAN} & 0.69 & 0.59 & 0.17 & 0.31\\
        \hline
        \ce{XGBoost} & 0.77 & 0.51 & 0.09 & 0.18\\
        \hline
        \multicolumn{4}{c}{Models on cross-validation} \\
        \hline
        \ce{Logistic Regression} & $0.70 \pm 0.01$ & $0.60 \pm 0.01$ & $0.12 \pm 0.01$ & $0.20 \pm 0.01$\\
        \hline
        \ce{Random Forest} & $0.76 \pm 0.01$ & $0.52 \pm 0.01$ & $0.11 \pm 0.01$ & $0.19 \pm 0.01$\\
        \hline
        \ce{CatBoost} & $0.75 \pm 0.01$ & $0.53 \pm 0.01$ & $0.12 \pm 0.01$ & $0.20 \pm 0.01$\\
        \hline
        \ce{TabPFN} & $0.82 \pm 0.01$ & $0.43 \pm 0.02$ & $0.09 \pm 0.01$ & $0.12 \pm 0.01$\\
        \hline
        \ce{KAN} & $0.67 \pm 0.01$ & $0.60 \pm 0.01$ & $0.19 \pm 0.02$ & $0.32 \pm 0.02$\\
        \hline
        \ce{XGBoost} & $0.76 \pm 0.01$ & $0.52 \pm 0.01$ & $0.12 \pm 0.01$ & $0.19 \pm 0.02$\\
        \hline
        \multicolumn{4}{c}{Models on external validation} \\
        \hline
        \ce{Logistic Regression}  & 0.73 & 0.57 & 0.15 & 0.12\\
        \hline
        \ce{Random Forest} & 0.74 & 0.53 & 0.15 & 0.11 \\
        \hline
        \ce{CatBoost}  & 0.74 & 0.54 & 0.15 & 0.11 \\
        \hline
        \ce{TabPFN} & 0.80 & 0.46 & 0.13 & 0.06 \\
        \hline
        \ce{KAN} & 0.66 & 0.61 & 0.22 & 0.26\\
        \hline
        \ce{XGBoost}  & 0.73 & 0.55 & 0.15 & 0.13 \\
    \bottomrule
    \end{tabular}
    \label{table:ext_test_prob_metr_feature_imp}

\end{table}

\section{External validation}
\textbf{Dataset description}  The key differences of this patient population from the original dataset were a lower ejection fraction (45.2\% vs. 56.2\%) lower incidence of cerebrovascular diseases (3.8\% vs. 12.3\%), previous PCI (3.8\% vs 41.3\%). At the same time, all these patients underwent ad-hoc PCI and they more often had anemia (18\% vs. 5\%).

\begin{table}[H]
    \caption{Classification metrics on external validation dataset}
    \label{validation}
    \centering
    \setlength{\tabcolsep}{8pt}
    \begin{tabular}{llllllllllll}
    \toprule
        {Model} & {AUC-ROC} & {F1-score} & {Precision} & {Recall} & {Accuracy}\\
        \midrule
        \multicolumn{6}{c}{Models without oversampling} \\
        \hline
        \ce{Logistic Regression} & 0.68 & 0.00 & 0.00 & 0.00 & \textbf{0.82}\\
        \hline
        \ce{Random Forest} & 0.65 & 0.00 & 0.00 & 0.00 & \textbf{0.82}\\
        \hline
        \ce{CatBoost} & 0.70 & 0.00 & 0.00 & 0.00 & \textbf{0.82}\\
        \hline
        \ce{TabPFN} & \textbf{0.71} & 0.00 & 0.00 & 0.00 & \textbf{0.82}\\
        \hline
        \ce{KAN} & 0.67 & \textbf{0.07} & \textbf{1.00} & \textbf{0.03} & \textbf{0.82}\\
        \hline
        \ce{XGBoost} & 0.68 & 0.00 & 0.00 & 0.00 & \textbf{0.82}\\
        \hline
        \multicolumn{6}{c}{Models with ARF oversampling} \\
        \hline
        \ce{Logistic Regression} & \textbf{0.69} & 0.00 & 0.00 & 0.00 & \textbf{0.82}\\
        \hline
        \ce{Random Forest} & 0.67 & 0.34 & 0.44 & 0.28 & 0.80\\
        \hline
        \ce{CatBoost} & 0.67 & 0.12 & 0.40 & 0.07 & 0.81\\
        \hline
        \ce{TabPFN} & 0.62 & 0.07 & \textbf{1.00} & 0.03 & \textbf{0.82}\\
        \hline
        \ce{KAN} & \textbf{0.69} & \textbf{0.50} & 0.39 & \textbf{0.69} & 0.75\\
        \hline
        \ce{XGBoost} & 0.64 & 0.11 & 0.25 & 0.07 & 0.79\\
        \hline
        \multicolumn{6}{c}{Models with TVAE oversampling} \\
        \hline
        \ce{Logistic Regression} & \textbf{0.69} & 0.00 & 0.00 & 0.00 & \textbf{0.82}\\
        \hline
        \ce{Random Forest} & 0.68 & 0.00 & 0.00 & 0.00 & 0.81\\
        \hline
        \ce{CatBoost} & \textbf{0.69} & 0.07 & \textbf{1.00} & 0.03 & \textbf{0.82}\\
        \hline
        \ce{TabPFN} & 0.63 & 0.00 & 0.00 & 0.00 & \textbf{0.82}\\
        \hline
        \ce{KAN} & 0.66 & \textbf{0.41} & 0.40 & \textbf{0.41} & 0.78\\
        \hline
        \ce{XGBoost} & 0.67 & 0.00 & 0.00 & 0.00 & \textbf{0.82}\\
        \hline
        \multicolumn{6}{c}{Models with GAN oversampling} \\
        \hline
        \ce{Logistic Regression} & \textbf{0.65} & 0.06 & 0.25 & 0.03 & 0.80\\
        \hline
        \ce{Random Forest} & 0.59 & 0.33 & 0.25 & 0.48 & 0.64\\
        \hline
        \ce{CatBoost}  & \textbf{0.65} & 0.27 & 0.24 & 0.31 & 0.70 \\
        \hline
        \ce{TabPFN} & 0.59 & 0.06 & \textbf{0.50} & 0.03 & \textbf{0.82}\\
        \hline
        \ce{KAN} & 0.60 & \textbf{0.36} & 0.26 & \textbf{0.55} & 0.63\\
        \hline
        \ce{XGBoost} & 0.64 & 0.13 & 0.18 & 0.10 & 0.75\\
        \hline
        \multicolumn{6}{c}{Models with Gaussian Copula oversampling} \\
        \hline
        \ce{Logistic Regression} & 0.69 & 0.00 & 0.00 & 0.00 & \textbf{0.82}\\
        \hline
        \ce{Random Forest} & \textbf{0.71} & 0.16 & \textbf{0.33} & 0.10 & 0.80\\
        \hline
        \ce{CatBoost} & 0.70 & 0.06 & 0.20 & 0.03 & 0.80 \\
        \hline
        \ce{TabPFN} & 0.61 & 0.00 & 0.00 & 0.00 & \textbf{0.82}\\
        \hline
        \ce{KAN} & 0.70 & \textbf{0.38} & \textbf{0.33} & \textbf{0.45} & 0.73 \\
        \hline
        \ce{XGBoost}  & 0.70 & 0.06 & 0.20 & 0.03 & 0.80\\
        \hline
        \multicolumn{6}{c}{Models with TabSyn oversampling} \\
        \hline
        \ce{Logistic Regression}  & 0.69 & 0.00 & 0.00 & 0.00 & 0.82 \\
        \hline
        \ce{Random Forest} & 0.67 & 0.00 & 0.00 & 0.00 & 0.82\\
        \hline
        \ce{CatBoost} & \textbf{0.71} & 0.00 & 0.00 & 0.00 & 0.82 \\
        \hline
        \ce{TabPFN} & 0.70 & 0.00 & 0.00 & 0.00 & 0.82\\
        \hline
        \ce{KAN} & \textbf{0.71} & \textbf{0.13} & \textbf{1.00} & \textbf{0.07} & \textbf{0.83}\\
        \hline
        \ce{XGBoost} & 0.69 & 0.00 & 0.00 & 0.00 & 0.82 \\
        \hline
        \multicolumn{6}{c}{Models with Edge cases oversampling} \\
        \hline
        \ce{Logistic Regression} & 0.68 & 0.00 & 0.00 & 0.00 & \textbf{0.82}\\
        \hline
        \ce{Random Forest} & 0.66 & \textbf{0.07} & \textbf{1.00} & \textbf{0.03} & \textbf{0.82}\\
        \hline
        \ce{CatBoost} & 0.69 & \textbf{0.07} & \textbf{1.00} & \textbf{0.03} & \textbf{0.82} \\
        \hline
        \ce{TabPFN} & \textbf{0.71} & 0.00 & 0.00 & 0.00 & \textbf{0.82}\\
        \hline
        \ce{KAN} & 0.68 & \textbf{0.07} & \textbf{1.00} & \textbf{0.03} & \textbf{0.82}\\
        \hline
        \ce{XGBoost}  & 0.68 & \textbf{0.07} & \textbf{1.00} & \textbf{0.03} & \textbf{0.82}\\
        
    \bottomrule
    \end{tabular}
    \label{table:ext_test_class_metr}

\end{table}

\begin{table}[H]
    \caption{Probabilistic metrics  on external validation dataset}
    \label{validation}
    \centering
    \setlength{\tabcolsep}{8pt}
    \begin{tabular}{lllll}
    \toprule
        {Model} & {Avg Confidence} & {Avg Entropy} & {Brier Score} & {ECE}\\
        \midrule
        \multicolumn{4}{c}{Models without oversampling} \\
        \hline
        \ce{Logistic Regression} & 0.93 & 0.26 & 0.16 & 0.11\\
        \hline
        \ce{Random Forest} & 0.93 & 0.25 & 0.15 & 0.11\\
        \hline
        \ce{CatBoost} & 0.93 & 0.25 & 0.15 & 0.11\\
        \hline
        \ce{TabPFN} & 0.93 & 0.22 & 0.15 & 0.12\\
        \hline
        \ce{KAN} & 0.90 & 0.27 & 0.14 & 0.08\\
        \hline
        \ce{XGBoost} & 0.93 & 0.25 & 0.15 & 0.11\\
        \hline
        \multicolumn{4}{c}{Models with ARF oversampling} \\
        \hline
        \ce{Logistic Regression} & 0.72 & 0.58 & 0.15 & 0.14\\
        \hline
        \ce{Random Forest} & 0.80 & 0.45 & 0.15 & 0.09\\
        \hline
        \ce{CatBoost} & 0.87 & 0.37 & 0.15 & 0.08\\
        \hline
        \ce{TabPFN} & 0.93 & 0.23 & 0.15 & 0.11\\
        \hline
        \ce{KAN} & 0.65 & 0.62 & 0.20 & 0.25\\
        \hline
        \ce{XGBoost} & 0.86 & 0.37 & 0.15 & 0.07\\
        \hline
        \multicolumn{4}{c}{Models with TVAE oversampling} \\
        \hline
        \ce{Logistic Regression}  & 0.73 & 0.58 & 0.15 & 0.12 \\
        \hline
        \ce{Random Forest}  & 0.91 & 0.25 & 0.14 & 0.10 \\
        \hline
        \ce{CatBoost} & 0.91 & 0.27 & 0.14 & 0.10\\
        \hline
        \ce{TabPFN} & 0.94 & 0.20 & 0.16 & 0.13\\
        \hline
        \ce{KAN} & 0.73 & 0.50 & 0.17 & 0.16\\
        \hline
        \ce{XGBoost}  & 0.91 & 0.26 & 0.15 & 0.10 \\
        \hline
        \multicolumn{4}{c}{Models with GAN oversampling} \\
        \hline
        \ce{Logistic Regression}  & 0.70 & 0.59 & 0.16 & 0.12\\
        \hline
        \ce{Random Forest} & 0.77 & 0.50 & 0.30 & 0.33 \\
        \hline
        \ce{CatBoost} & 0.80 & 0.45 & 0.21 & 0.16 \\
        \hline
        \ce{TabPFN} & 0.90 & 0.29 & 0.15 & 0.10\\
        \hline
        \ce{KAN} & 0.78 & 0.47 & 0.28 & 0.29\\
        \hline
        \ce{XGBoost} & 0.79 & 0.47 & 0.17 & 0.10\\
        \hline
        \multicolumn{4}{c}{Models with Gaussian Copula oversampling} \\
        \hline
        \ce{Logistic Regression} & 0.72 & 0.59 & 0.15 & 0.12\\
        \hline
        \ce{Random Forest} & 0.78 & 0.50 & 0.15 & 0.08\\
        \hline
        \ce{CatBoost} & 0.88 & 0.35 & 0.15 & 0.09\\
        \hline
        \ce{TabPFN} & 0.94 & 0.21 & 0.16 & 0.13\\
        \hline
        \ce{KAN} & 0.61 & 0.66 & 0.20 & 0.26\\
        \hline
        \ce{XGBoost}  & 0.87 & 0.37 & 0.15 & 0.08\\
        \hline
        \multicolumn{4}{c}{Models with TabSyn oversampling} \\
        \hline
        \ce{Logistic Regression}  & 0.84 & 0.43 & 0.14 & 0.03 \\
        \hline
        \ce{Random Forest} & 0.93 & 0.25 & 0.16 & 0.11\\
        \hline
        \ce{CatBoost} & 0.92 & 0.27 & 0.15 & 0.11 \\
        \hline
        \ce{TabPFN} & 0.94 & 0.22 & 0.15 & 0.12\\
        \hline
        \ce{KAN} & 0.88 & 0.29 & 0.14 & 0.08\\
        \hline
        \ce{XGBoost} & 0.92 & 0.26 & 0.15 & 0.11 \\
        \hline
        \multicolumn{4}{c}{Models with Edge cases oversampling} \\
        \hline
        \ce{Logistic Regression} & 0.92 & 0.27 & 0.16 & 0.10\\
        \hline
        \ce{Random Forest} & 0.93 & 0.26 & 0.15 & 0.11\\
        \hline
        \ce{CatBoost}  & 0.93 & 0.26 & 0.15 & 0.11 \\
        \hline
        \ce{TabPFN} & 0.95 & 0.20 & 0.16 & 0.13\\
        \hline
        \ce{KAN} & 0.90 & 0.28 & 0.14 & 0.08\\
        \hline
        \ce{XGBoost} & 0.92 & 0.27 & 0.15 & 0.10\\
        
    \bottomrule
    \end{tabular}
    \label{table:ext_test_prob_metr}

\end{table}

\textbf{Performance.} On the external dataset (Tables ~\ref{table:ext_test_class_metr} and ~\ref{table:ext_test_prob_metr}), a decline in the target metric AUC-ROC to the range of 0.59–0.71 is observed across all models, which can be attributed to the small sample size. However, external validation also reveals dependencies consistent with those identified in the original dataset. The patterns observed in the original data are thus confirmed in the external validation. Models trained without synthetic data tend to avoid predicting death probabilities above 50\%, which results in zero precision and recall, except for KAN, where recall remains very low. Models trained with synthetic data generated by ARF and GAN exhibit an improvement in the F1 score relative to models without synthetic data, consistent with the findings from the experiments on the original dataset. In contrast, models trained with other types of synthetic data do not yield substantial improvements in performance metrics on external data.

\section{Discussion}
The main findings of the present study were as follows. First, without addressing class imbalance, models appear well-calibrated and achieve high accuracy but completely fail to identify the minority class, making them unsuitable for practical use. Second, oversampling does not simply improve traditional performance metrics but fundamentally alters model behavior. ARF and GAN oversampling are the most effective approaches, substantially improving Recall and F1 at the cost of reduced accuracy. Generative approaches such as TVAE and Gaussian Copula are less effective and often detrimental. Third, inclusion of synthetic data tends to reduce the average confidence of the models. Fourth, there is a clear trade-off between calibration and sensitivity: models such as TabPFN maintain excellent calibration but fail to identify minority-class samples, whereas models such as KAN sacrifice calibration and accuracy in favor of detecting a higher proportion of positive cases. Fifth, this study identified 4 clinical factors: age, ejection fraction, peripheral artery disease and cerebrovascular disease as the most important features. Sixth, developed prediction model demonstrated good metrics in external validation in patients with acute myocardial infarction.

Reliable risk prediction long-term outcomes depends on models that remain accurate and well-calibrated precisely for rare, clinically severe patients. Yet class imbalance and distribution shift routinely mask failures. Motivated by our observation that models with excellent test-set scores produced nonsensical outputs once severe, previously unseen patients were introduced, we ask a simple question: can adding obvious, extreme-but-plausible minority-class profiles and in-distribution synthetic data make tabular risk models both perform better and behave sensibly on these very cases? 
Synthetic data generation offers perspective solution to class imbalance, which is very common in medicine due to low incidence of adverse events. Recently, Zawadzki et al., used SDG to augment training data for Catboost in classifying chronic heart failure\cite{zawadzki2023synthetic}. They found that utilizing SDG modestly improved classification performance compared to a baseline model.

Results of our study demonstrates that different synthetic data generation methods have distinct effects on model performance. ARF enhances sensitivity to the minority class while largely preserving ranking ability within the dataset. TabSyn, although promising in large-scale settings, fails to capture the minority class distribution in small, imbalanced datasets. GAN proves effective for augmenting the minority class, leading to consistent improvements in both test and cross-validation performance. In contrast, TVAE and Gaussian Copula produce only moderate results for this task.

Also, in this study, we tested an approach to address class imbalance by augmenting the training set with data generated using ARF (N = 500) and edge cases (N = 500). The results of this experiment are presented in Tables ~\ref{table:arf_edge_class_metr} and ~\ref{table:arf_edge_prob_metr}. This analysis was conducted to assess the extent to which combining different methods can improve the metrics on the test set.

Several recent studies have demonstrated the feasibility of predicting mortality after PCI using machine learning algorithms. For example, Burrello et al. developed a predictive model for post-PCI mortality that achieved clinically relevant accuracy using a cohort of more than 2,000 patients \cite{burrello2022prediction}. Similarly, Kang et al. integrated data from two large bifurcation registries (RAIN and COBIS) and identified both clinical and lesion-specific predictors of adverse cardiovascular events, though the discriminatory ability of the final model remained moderate (AUC ~0.657) \cite{kang2022impact}. An important limitation of this study is the small number of variables used for analysis. Initially, only 18 variables were included. 
While our study initially included 283 features. Importantly, patient characteristics and procedural characteristics vary significantly across populations, which highlights the need for further external validation and adaptation of predictive models to diverse clinical cohorts.
External validation is a critical step in the development and evaluation of prognostic machine learning models in medicine. While internal validation techniques such as cross-validation or random train–test splits are useful for estimating model performance on unseen subsets of the same dataset, they cannot fully assess the model’s ability to generalize to independent patient populations. Models trained and validated solely on a single cohort are at high risk of overfitting to cohort-specific patterns such as site-dependent treatment practices, measurement procedures, or demographic distributions.
By contrast, external validation tests the model on data that were not used at any stage of model training or parameter selection and that ideally originate from a different institution, geographic region, or time period. This process allows investigators to evaluate the robustness of the predictive signal, to detect over-optimistic performance estimates, and to ensure that the model captures clinically meaningful associations rather than spurious correlations. Our results confirm that developed prediction model can be used to evaluate risk of cardiac mortality in patients with various forms of coronary artery disease. This is due to that the key factors that have the greatest impact on the prognosis are clinical (age, anemia, cerebrovascular diseases, peripheral disease, the presence of aortic stenosis, etc.), while angiographic and procedural factors have less impact on the prognosis in patients. These data are consistent with the results of the study of Kang et al, in which clinical factors had a greater impact on all-cause death and myocardial infarction, whereas lesion- specific factors were predominantly associated with lesion-oriented clinical outcomes\cite{kang2022impact}. 

In addition to standard classification metrics, probabilistic measures were used to assess the reliability of predicted probabilities, which is especially important with class imbalance. Average Confidence reflects model overconfidence, while Average Entropy captures uncertainty; low entropy may suggest bias toward the majority class. The Brier score evaluates overall calibration by comparing predicted probabilities to true outcomes, and the Expected Calibration Error (ECE) measures the gap between confidence and accuracy across bins. Together, these metrics complement discrimination-focused measures by showing how well probability estimates match real outcome frequencies—crucial in clinical risk prediction, where calibrated probabilities guide decision-making more effectively than raw classifications.

In this study, although models trained without synthetic data demonstrated strong performance in terms of AUC-ROC, they assigned excessively low probabilities to positive instances, as reflected by the F1-score and low Precision. At the same time, models trained without synthetic data exhibited high confidence in their predictions, which is evident from the Average Confidence metrics. The introduction of synthetic data reduces this measure, leading to less categorical predictions. In other words, synthetic data aids in the detection of positive-class instances in this task. It is also worth noting the low average probability entropy observed in models without synthetic data, which is likewise alleviated when the training set is augmented with synthetic samples.

To evaluate the contribution of individual features, permutation-based feature importance was applied, with the average impact measured over multiple iterations. This method was selected for its universality, allowing consistent assessment across all models. The analysis provides insights into which features most strongly influence predictions, though further research is needed to assess whether synthetic data accurately preserves these dependencies in clinical contexts.

\textbf{Study limitations.} 
Before interpreting the results of this study, it is important to acknowledge several limitations of our study. First, the datasets were relatively small, heterogeneous, and exhibited class imbalance, which presents challenges for both model training and reliable performance assessment. Additionally, missing values and potential biases in the original data may affect the generalizability of the models.

Despite these challenges, the experimental results indicate that generative approaches, including ARF, CTGAN, TVAE, Copula, and TabSyn, can enhance classification performance in imbalanced tabular datasets. These methods demonstrate the ability to capture complex feature dependencies and improve predictive accuracy when compared to traditional oversampling techniques.

Second, the models were trained and evaluated on retrospective datasets, which limits conclusions about their prospective clinical applicability. Third, the fidelity and interpretability of synthetic data generated by these models require more in-depth analysis to ensure that generated samples do not introduce unintended biases or distortions. Third, limitation of this study is that only 500 positive-class samples were added to the training set. This decision was motivated by the need to evaluate the contribution of different oversampling methods rather than to achieve the best possible performance metrics for each method. This hyperparameter can be determined using the validation set or cross-validation through hyperparameter search methods. However, such an experiment was not conducted in the present study.

\begin{table}[H]
    \caption{Classification metrics on test and cross-validation with Edge cases and ARF oversampling}
    \label{validation}
    \centering
    \setlength{\tabcolsep}{10pt}
    \begin{tabular}{llllllllllll}
    \toprule
        {Model} & {AUC-ROC} & {F1-score} & {Precision} & {Recall} & {Accuracy}\\
        \midrule
        \multicolumn{6}{c}{Models with EDGE and ARF oversampling on test set} \\
        \hline
        \ce{Logistic Regression} & 0.78 & 0.26 & 0.40 & 0.19 & 0.92\\
        \hline
        \ce{Random Forest} & 0.78 & 0.27 & 0.26 & 0.29 & 0.88\\
        \hline
        \ce{CatBoost} & 0.76 & 0.14 & 0.25 & 0.10 & 0.91\\
        \hline
        \ce{TabPFN} & 0.71 & 0.05 & 0.17 & 0.03 & 0.91\\
        \hline
        \ce{KAN} & 0.76 & 0.30 & 0.20 & 0.65 & 0.77\\
        \hline
        \ce{XGBoost} & 0.78 & 0.20 & 0.28 & 0.16 & 0.90\\
        \hline
        \multicolumn{6}{c}{Models with EDGE and ARF oversampling on cross-validation} \\
        \hline
        \ce{Logistic Regression} & $0.72 \pm 0.11$ & $0.16 \pm 0.15$ & $0.23 \pm 0.24$ & $0.13 \pm 0.12$ & $0.90 \pm 0.02$\\
        \hline
        \ce{Random Forest} & $0.71 \pm 0.11$ & $0.23 \pm 0.15$ & $0.19 \pm 0.12$ & $0.29 \pm 0.21$ & $0.85 \pm 0.03$\\
        \hline
        \ce{CatBoost} & $0.71 \pm 0.10$ & $0.18 \pm 0.15$ & $0.21 \pm 0.17$ & $0.16 \pm 0.14$ & $0.89 \pm 0.02$\\
        \hline
        \ce{TabPFN} & $0.65 \pm 0.11$ & $0.06 \pm 0.07$ & $0.09 \pm 0.11$ & $0.04 \pm 0.05$ & $0.90 \pm 0.01$\\
        \hline
        \ce{KAN} & $0.72 \pm 0.11$ & $0.24 \pm 0.07$ & $0.16 \pm 0.07$ & $0.55 \pm 0.18$ & $0.73 \pm 0.04$\\
        \hline
        \ce{XGBoost} & $0.70 \pm 0.11$ & $0.20 \pm 0.17$ & $0.19 \pm 0.16$ & $0.22 \pm 0.20$ & $0.87 \pm 0.03$\\
    \bottomrule
    \end{tabular}
    \label{table:arf_edge_class_metr}

\end{table}

\begin{table}[H]
    \caption{Probability metrics on test and cross-validation with Edge cases and ARF oversampling}
    \label{validation}
    \centering
    \setlength{\tabcolsep}{10pt}
    \begin{tabular}{lllll}
    \toprule
        {Model} & {Avg Confidence} & {Avg Entropy} & {Brier Score} & {ECE}\\
        \midrule
        \multicolumn{4}{c}{Models with EDGE and ARF oversampling on test set} \\
        \hline
        \ce{Logistic Regression} & 0.72 & 0.58 & 0.11 & 0.21\\
        \hline
        \ce{Random Forest} & 0.81 & 0.42 & 0.09 & 0.13\\
        \hline
        \ce{CatBoost} & 0.89 & 0.32 & 0.07 & 0.05\\
        \hline
        \ce{TabPFN} & 0.93 & 0.22 & 0.07 & 0.02\\
        \hline
        \ce{KAN} & 0.69 & 0.60 & 0.18 & 0.31\\
        \hline
        \ce{XGBoost} & 0.84 & 0.40 & 0.08 & 0.10\\
        \hline
        \multicolumn{4}{c}{Models with EDGE and ARF oversampling on cross-validation} \\
        \hline
        \ce{Logistic Regression} & $0.70 \pm 0.01$ & $0.59 \pm 0.01$ & $0.12 \pm 0.01$ & $0.23 \pm 0.01$\\
        \hline
        \ce{Random Forest} & $0.81 \pm 0.01$ & $0.44 \pm 0.02$ & $0.11 \pm 0.02$ & $0.15 \pm 0.02$\\
        \hline
        \ce{CatBoost} & $0.88 \pm 0.01$ & $0.32 \pm 0.01$ & $0.09 \pm 0.02$ & $0.08 \pm 0.02$\\
        \hline
        \ce{TabPFN} & $0.93 \pm 0.00$ & $0.23 \pm 0.01$ & $0.08 \pm 0.01$ & $0.05 \pm 0.01$\\
        \hline
        \ce{KAN} & $0.68 \pm 0.01$ & $0.60 \pm 0.01$ & $0.19 \pm 0.02$ & $0.33 \pm 0.02$\\
        \hline
        \ce{XGBoost} & $0.84 \pm 0.01$ & $0.40 \pm 0.02$ & $0.10 \pm 0.02$ & $0.12 \pm 0.02$\\
    \bottomrule
    \end{tabular}
    \label{table:arf_edge_prob_metr}

\end{table}

\textbf{Key findings.} In mortality prediction tasks, it has been observed that models tend to underestimate the probability of death under conditions of class imbalance. This effect is particularly pronounced in edge cases, where clinical values indicate life-threatening conditions. In such situations, one would expect the models to assign high probabilities of mortality; however, their behavior remains inadequate. The use of synthetic data has been shown to mitigate this issue, improving both the calibration of predicted probabilities and the overall classification metrics.

Evaluating models on both edge cases and real-world clinical data provides a more comprehensive assessment of their ability to extrapolate learned patterns and to capture critical regions in the feature space. This dual testing strategy highlights whether models can meaningfully recognize and respond to scenarios of high clinical risk.

Furthermore, feature analysis has revealed that clinical variables, along with selected non-clinical predictors, contribute substantially to the models’ predictive performance. Notably, restricting the feature set to the most informative variables results in a considerable improvement in performance metrics. This observation suggests that noise in the data plays a significant role in limiting the effectiveness of predictive models, and that careful feature selection or dimensionality reduction may enhance their robustness.

\textbf{Clinical implications.} Developed model allows us to identify a group of patients with a high risk of developing adverse events and carry out more intensive treatment and preventive measures in relation to modifiable risk factors.

\textbf{Future work.} Further expansion of the dataset and prospective validation on patients with various forms of coronary artery disease may allow the implementation of this model in clinical practice.

\section{Conclusion}

In this study, we developed predictive models for assessing three-year cardiac mortality following  PCI. The results demonstrate that the choice of generative model substantially affects classification performance across evaluation metrics. While ARF and GAN-based augmentation improved minority class detection and enhanced model sensitivity, other methods such as TVAE, Gaussian Copula, and TabSyn yielded more moderate or inconsistent gains. Additionally, edge-case synthetic data provided further insights into the models’ robustness in clinically critical scenarios. Overall, our findings highlight both the potential and the limitations of synthetic data in improving clinical risk prediction when working with constrained datasets.

\printbibliography

\section{Declarations}

\textbf{Author Contributions}. Daniil Burakov, HSE University, Moscow, Russian Federation — code and data analysis; drafting of the manuscript; critical revision; final approval.

Ivan Petrov, HSE University, Moscow, Russian Federation — proof of concept and preliminary results; drafting; critical revision; final approval.

Dmitrii Khelimskii, MD, PhD, Meshalkin National Medical Research Center, Ministry of Health of Russian Federation, Novosibirsk, Russian Federation — conception and design; data collection; drafting; critical revision; final approval.

Ivan Bessonov, MD, PhD, Tyumen Cardiology Research Center, Tomsk National Research Medical Center, Russian Academy of Sciences, Tomsk 625026, Russian Federation — external validation; drafting; critical revision.

Mikhail Lazarev, HSE University, Moscow, Russian Federation — supervision; conception and design; drafting; critical revision; final approval; funding.

\textbf{Competing Interests}. The authors declare that they have no known competing financial interests or personal relationships that could have appeared to influence the work reported in this paper.

\textbf{Ethics Statement}. This retrospective study was conducted in accordance with the Declaration of Helsinki and was approved by the E.N. Meshalkin National Medical Research Center review board (protocol number 18 from 08.12.2017). The ethics committee granted a waiver of informed consent as the study involved the analysis of existing anonymized data and posed minimal risk to the participants.

\textbf{Funding}. The work was supported by the grant for research centers in the field of AI provided by the Ministry of Economic Development of the Russian Federation in accordance with the agreement 000000С313925P4E0002 and the agreement with HSE University № 139-15-2025-009.

\section{Appendix}

 \textbf{Evaluation Metrics and Results}: The Tables ~\ref{table:test_class_metr_full} and ~\ref{table:test_prob_metr_full} present the complete experimental results obtained on the test set with clean data as well as with synthetic data generated using methods such as ARF, TVAE, CTGAN, Gaussian Copula, TabSyn, and with the inclusion of edge cases. These tables report the main classification metrics and the probability characteristics produced by the models on the test dataset. The best-performing metrics are highlighted for each class of oversampling mitigation methods. In addition, to illustrate the distribution of the metrics, the Tables ~\ref{table:cv_class_metr_full} and ~\ref{table:cv_prob_metr_full} provide the mean values and standard deviations computed via ten-fold cross-validation.

\begin{table}[H]
    \caption{Classification metrics on test set and probability distribution on edge cases}
    \label{validation}
    \centering
    \setlength{\tabcolsep}{4pt}
    \begin{tabular}{llllllllllll}
    \toprule
        {Model} & {AUC-ROC} & {F1-score} & {Precision} & {Recall} & {Accuracy} & {Q0} & {Q50} & {Q99} & {Mean} & {Std} \\
        \midrule
        \multicolumn{6}{c}{Models without oversampling} \\
        \hline
        \ce{Logistic Regression} & 0.78 & 0 & 0 &  0 & 0.92 &  0.16 & 0.23 & 0.30 & 0.23 & 0.03\\
        \hline
        \ce{Random Forest} & 0.78 & 0 & 0 &  0 & 0.92 & 0.34 & 0.51 & 0.58 & 0.51 & 0.04\\
        \hline
        \ce{CatBoost} & 0.76 & 0 & 0 & 0 & 0.92 &  0.33 & 0.63 & 0.74 & 0.61 & 0.07\\
        \hline
        \ce{TabPFN} & \textbf{0.82} & 0 & 0 & 0 & 0.92 & 0.46 & 0.65 & 0.72 & 0.64 & 0.04\\
        \hline
        \ce{KAN} & 0.79 & \textbf{0.06} & \textbf{1} & \textbf{0.03} & \textbf{0.93} & 0.68 & 0.99 & 0.99 & 0.98 & 0.03\\
        \hline
        \ce{XGBoost} & 0.79 & \textbf{0.06} & \textbf{1} & \textbf{0.03} & \textbf{0.93}  & 0.40 & 0.64 & 0.74 & 0.62 & 0.06\\
        \hline
        \multicolumn{6}{c}{Models with ARF oversampling} \\
        \hline
        \ce{Logistic Regression} & 0.77 & 0.06 & 0.5 &  0.03 & \textbf{0.92}  & 0.53 & 0.71 & 0.81 & 0.71 & 0.05 \\
        \hline
        \ce{Random Forest} & \textbf{0.78} & \textbf{0.29} & 0.28 & 0.29 & 0.89 & 0.74 & 0.88 & 0.93 & 0.88 & 0.03\\
        \hline
        \ce{CatBoost} & 0.77 & 0.14 & 0.23 & 0.1 & 0.90 & 0.83 & 0.96 & 0.97 & 0.95 & 0.02\\
        \hline
        \ce{TabPFN} & 0.71 & 0.05 & 0.17 & 0.03 & 0.91 & 0.75 & 0.97 & 0.99 & 0.95 & 0.04\\
        \hline
        \ce{KAN} & 0.75 & 0.27 & 0.18 & \textbf{0.55} & 0.78 & 0.56 & 0.99 & 0.99 & 0.96 & 0.05\\
        \hline
        \ce{XGBoost} & 0.75 & 0.19 & \textbf{0.33} & 0.13 & 0.91 & 0.66 & 0.96 & 0.98 & 0.94 & 0.04\\
        \hline
        \multicolumn{6}{c}{Models with TVAE oversampling} \\
        \hline
        \ce{Logistic Regression} & \textbf{0.73} & 0 & 0 & 0 & \textbf{0.92}  & 0.30 & 0.40 & 0.50 & 0.40 & 0.04\\
        \hline
        \ce{Random Forest} & 0.66 & \textbf{0.16} & \textbf{0.21} & 0.13 & 0.9 & 0.26 & 0.80 & 0.91 & 0.71 & 0.16\\
        \hline
        \ce{CatBoost} & 0.68 & 0.11 & 0.4 & 0.06 & \textbf{0.92} & 0.58 & 0.90 & 0.97 & 0.87 & 0.08\\
        \hline
        \ce{TabPFN} & 0.64 & 0 & 0 & 0 & \textbf{0.92} & 0.20 & 0.52 & 0.95 & 0.55 & 0.15\\
        \hline
        \ce{KAN} & 0.58 & \textbf{0.16} & 0.1 & \textbf{0.39} & 0.69 & 0.00 & 0.13 & 0.98 & 0.31 & 0.34\\
        \hline
        \ce{XGBoost} & 0.63 & 0.06 & 0.2 & 0.03 & \textbf{0.92} & 0.46 & 0.90 & 0.96 & 0.80 & 0.16\\
        \hline
        \multicolumn{6}{c}{Models with GAN oversampling} \\
        \hline
        \ce{Logistic Regression} & 0.76 & 0.11 & 0.29 & 0.06 & \textbf{0.92} & 0.48 & 0.80 & 0.89 & 0.78 & 0.08\\
        \hline
        \ce{Random Forest} & \textbf{0.78} & 0.29 & 0.32 & 0.26 & 0.90 & 0.73 & 0.88 & 0.93 & 0.88 & 0.02\\
        \hline
        \ce{CatBoost} & 0.77 & 0.28 & \textbf{0.37} & 0.23 & 0.91 & 0.68 & 0.83 & 0.92 & 0.83 & 0.05\\
        \hline
        \ce{TabPFN} & 0.76 & 0.06 & 0.2 & 0.03 & \textbf{0.92} & 0.91 & 0.98 & 0.99 & 0.98 & 0.01\\
        \hline
        \ce{KAN} & 0.76 & \textbf{0.30} & 0.25 & \textbf{0.39} & 0.87 & 0.06 & 0.99 & 0.99 & 0.94 & 0.13\\
        \hline
        \ce{XGBoost} & 0.75 & 0.14 & 0.27 & 0.1 & 0.91 & 0.61 & 0.95 & 0.98 & 0.93 & 0.05\\
        \hline
        \multicolumn{6}{c}{Models with Gaussian Copula oversampling} \\
        \hline
        \ce{Logistic Regression} & 0.76 & 0 & 0 & 0 & \textbf{0.92} & 0.46 & 0.58 & 0.66 & 0.58 & 0.04\\
        \hline
        \ce{Random Forest} & \textbf{0.78} & 0.2 & \textbf{0.2} & 0.19 & 0.88 &  0.72 & 0.86 & 0.95 & 0.86 & 0.05\\
        \hline
        \ce{CatBoost} & 0.77 & 0.05 & 0.11 & 0.03 & 0.91 & 0.37 & 0.72 & 0.98 & 0.76 & 0.17\\
        \hline
        \ce{TabPFN} & 0.71 & 0.05 & 0.17 & 0.03 & 0.91 & 0.56 & 0.89 & 0.99 & 0.88 & 0.10\\
        \hline
        \ce{KAN} & 0.73 & \textbf{0.27} & 0.18 & \textbf{0.58} & 0.76 & 0.26 & 0.90 & 0.97 & 0.87 & 0.11\\
        \hline
        \ce{XGBoost} & 0.73 & 0.05 & 0.13 & 0.03 & 0.91 & 0.42 & 0.72 & 0.99 & 0.76 & 0.18\\
        \hline
        \multicolumn{6}{c}{Models with TabSyn oversampling} \\
        \hline
        \ce{Logistic Regression} & 0.74 & 0.07 & 0.08 & 0.06 & 0.87 & 0.37 & 0.86 & 0.91 & 0.78 & 0.16\\
        \hline
        \ce{Random Forest} & 0.75 & 0.09 & 0.13 & 0.06 & 0.88 & 0.28 & 0.84 & 0.90 & 0.76 & 0.18\\
        \hline
        \ce{CatBoost} & 0.75 & 0.08 & 0.11 & 0.06 & \textbf{0.89} & 0.19 & 0.89 & 0.92 & 0.80 & 0.19\\
        \hline
        \ce{TabPFN} & \textbf{0.76} & 0.08 & 0.11 & 0.06 & \textbf{0.89} & 0.27 & 0.88 & 0.92 & 0.79 & 0.18\\
        \hline
        \ce{KAN} & 0.72 & \textbf{0.21} & \textbf{0.18} & \textbf{0.25} & 0.85 & 0.27 & 0.88 & 0.92 & 0.79 & 0.18\\
        \hline
        \ce{XGBoost} & 0.75 & 0.07 & 0.1 & 0.06 & \textbf{0.89} & 0.27 & 0.89 & 0.92 & 0.79 & 0.18\\
        \hline
        \multicolumn{6}{c}{Models with Edge cases oversampling} \\
        \hline
        \ce{Logistic Regression} & 0.78 & 0 & 0 & 0 & 0.92 & 0.60 & 0.97 & 0.99 & 0.95 & 0.05\\
        \hline
        \ce{Random Forest} & 0.78 & 0 & 0 & 0 & 0.92 &  0.98 & 0.99 & 0.99 & 0.99 & 0.00\\
        \hline
        \ce{CatBoost} & 0.77 & 0.12 & 0.67 & 0.06 & \textbf{0.93} & 0.98 & 0.98 & 0.98 & 0.98 & 0.00\\
        \hline
        \ce{TabPFN} & \textbf{0.81} & 0.06 & \textbf{1} & 0.03 & \textbf{0.93} & 0.99 & 0.99 & 0.99 & 0.99 & 0.00\\
        \hline
        \ce{KAN} & 0.76 & \textbf{0.14} & 0.27 & \textbf{0.1} & 0.91 & 0.92 & 0.99 & 0.99 & 0.99 & 0.01\\
        \hline
        \ce{XGBoost} & 0.79 & 0.06 & 0.5 & 0.03 & 0.92 & 0.99 & 0.99 & 0.99 & 0.99 & 0.00\\
        
    \bottomrule
    \end{tabular}
    \label{table:test_class_metr_full}

\end{table}

\begin{table}[H]
    \caption{Probabilistic metrics on test set}
    \label{validation}
    \centering
    \setlength{\tabcolsep}{6pt}
    \begin{tabular}{llllll}
    \toprule
        {Model} & {Avg Confidence} & {Avg Entropy} & {Brier score} & {ECE}\\
        \midrule
        \multicolumn{4}{c}{Models without oversampling} \\
        \hline
        \ce{Logistic Regression} & 0.92 & 0.27 & 0.07 & 0.03 \\
        \hline
        \ce{Random Forest} & 0.92 & 0.26 & 0.06 & 0.03\\
        \hline
        \ce{CatBoost} & 0.92 & 0.27 & 0.06 & 0.02\\
        \hline
        \ce{TabPFN} & 0.93 & 0.22 & 0.06 & 0.03\\
        \hline
        \ce{KAN} & 0.92 & 0.26 & 0.06 & 0.04\\
        \hline
        \ce{XGBoost} & 0.92 & 0.26 & 0.06 & 0.04\\
        \hline
        \multicolumn{4}{c}{Models with ARF oversampling} \\
        \hline
        \ce{Logistic Regression} & 0.72 & 0.59 & 0.11 & 0.21 \\
        \hline
        \ce{Random Forest} & 0.83 & 0.40 & 0.09 & 0.11\\
        \hline
        \ce{CatBoost} & 0.88 & 0.33 & 0.07 & 0.06\\
        \hline
        \ce{TabPFN} & 0.93 & 0.24 & 0.07 & 0.02\\
        \hline
        \ce{KAN} & 0.69 & 0.6 & 0.18 & 0.31\\ 
        \hline
        \ce{XGBoost} & 0.88 & 0.34 & 0.07 & 0.06\\
        \hline
        \multicolumn{4}{c}{Models with TVAE oversampling} \\
        \hline
        \ce{Logistic Regression} & 0.71 & 0.6 & 0.11 & 0.22 \\
        \hline
        \ce{Random Forest} & 0.85 & 0.35 & 0.08 & 0.08\\
        \hline
        \ce{CatBoost} & 0.89 & 0.31 & 0.07 & 0.04\\
        \hline
        \ce{TabPFN} & 0.93 & 0.23 & 0.07 & 0.02\\
        \hline
        \ce{KAN} & 0.70 & 0.56 & 0.2 & 0.31\\ 
        \hline
        \ce{XGBoost} & 0.89 & 0.32 & 0.08 & 0.06\\
        \hline
        \multicolumn{4}{c}{Models with GAN oversampling} \\
        \hline
        \ce{Logistic Regression} & 0.76 & 0.53 & 0.09 & 0.17 \\
        \hline
        \ce{Random Forest} & 0.85 & 0.38 & 0.08 & 0.09\\
        \hline
        \ce{CatBoost} & 0.89 & 0.31 & 0.07 & 0.06\\
        \hline
        \ce{TabPFN} & 0.93 & 0.24 & 0.07 & 0.01\\
        \hline
        \ce{KAN} & 0.82 & 0.41 & 0.1 & 0.15\\ 
        \hline
        \ce{XGBoost} & 0.89 & 0.31 & 0.07 & 0.05\\
        \hline
        \multicolumn{4}{c}{Models with Gaussian Copula oversampling} \\
        \hline
        \ce{Logistic Regression} & 0.71 & 0.60 & 0.11 & 0.22 \\
        \hline
        \ce{Random Forest} &  0.82 & 0.43 & 0.09 & 0.12\\
        \hline
        \ce{CatBoost} & 0.89 & 0.32 & 0.07 & 0.05\\
        \hline
        \ce{TabPFN} & 0.93 & 0.22 & 0.08 & 0.03\\
        \hline
        \ce{KAN} & 0.77 & 0.49 & 0.13 & 0.20\\ 
        \hline
        \ce{XGBoost} & 0.89 & 0.33 & 0.07 & 0.05\\
        \hline
        \multicolumn{4}{c}{Models with TabSyn oversampling} \\
        \hline
        \ce{Logistic Regression} & 0.82 & 0.46 & 0.09 & 0.12 \\
        \hline
        \ce{Random Forest} &  0.90 & 0.28 & 0.09 & 0.06\\
        \hline
        \ce{CatBoost} & 0.91 & 0.29 & 0.09 & 0.07 \\
        \hline
        \ce{TabPFN} & 0.92 & 0.24 & 0.09 & 0.05\\
        \hline
        \ce{KAN} & 0.86 & 0.35 & 0.11 & 0.12\\ 
        \hline
        \ce{XGBoost} & 0.90 & 0.28 & 0.09 & 0.07\\
        \hline
        \multicolumn{4}{c}{Models with Edge cases oversampling} \\
        \hline
        \ce{Logistic Regression} & 0.91 & 0.29 & 0.06 & 0.04 \\
        \hline
        \ce{Random Forest} &  0.92 & 0.27 & 0.06 & 0.04\\
        \hline
        \ce{CatBoost} & 0.92 & 0.26 & 0.06 & 0.03 \\
        \hline
        \ce{TabPFN} & 0.94 & 0.2 & 0.06 & 0.03\\
        \hline
        \ce{KAN} & 0.88 & 0.32 & 0.07 & 0.05\\ 
        \hline
        \ce{XGBoost} & 0.92 & 0.27 & 0.06 & 0.05\\
    \bottomrule
    \end{tabular}
    \label{table:test_prob_metr_full}

\end{table}

\begin{table}[H]
    \caption{Classification metrics on Cross validation}
    \label{validation}
    \centering
    \setlength{\tabcolsep}{10pt}
    \begin{tabular}{llllll}
    \toprule
        {Model} & {AUC-ROC} & {F1-score} & {Precision} & {Recall} & {Accuracy} \\
        \midrule
        \multicolumn{6}{c}{Models without oversampling} \\
        \hline
        \ce{Logistic Regression} & $\textbf{0.72} \pm 0.11$ & $0.00 \pm 0.00$ & $0.00 \pm 0.00$ & $0.00 \pm 0.00$ & $0.92 \pm 0.00$ \\
        \hline
        \ce{Random Forest} & $\textbf{0.72} \pm 0.11$ & $0.00 \pm 0.00$ &  $0.00 \pm 0.00$ &  $0.00 \pm 0.00$ & $0.92 \pm 0.00$ \\
        \hline
        \ce{CatBoost} & $0.70 \pm 0.11$ & $\textbf{0.03} \pm 0.06$ & $0.15 \pm 0.34$ & $\textbf{0.02} \pm 0.03$ & $0.92 \pm 0.00$ \\
        \hline
        \ce{TabPFN} & $0.70 \pm 0.12$ & $0.00 \pm 0.00$ & $0.00 \pm 0.00$ & $0.00 \pm 0.00$ & $0.92 \pm 0.00$ \\
        \hline
        \ce{KAN} & $\textbf{0.72} \pm 0.12$ & $\textbf{0.03} \pm 0.06$ & $\textbf{0.2} \pm 0.42$ & $\textbf{0.02} \pm 0.03$ & $0.92 \pm 0.01$ \\
        \hline
        \ce{XGBoost} & $\textbf{0.72} \pm 0.12$ & $\textbf{0.03} \pm 0.06$ & $\textbf{0.2} \pm 0.42$ & $\textbf{0.02} \pm 0.03$ & $0.92 \pm 0.00$ \\
        \hline
        \multicolumn{6}{c}{Models with ARF oversampling} \\
        \hline
        \ce{Logistic Regression} & $\textbf{0.72} \pm 0.11$ & $0.09 \pm 0.12$ & $\textbf{0.26} \pm 0.35$ & $0.06 \pm 0.07$ & $\textbf{0.92} \pm 0.01$ \\
        \hline
        \ce{Random Forest} & $0.71 \pm 0.10$ & $0.22 \pm 0.15$ & $0.18 \pm 0.12$ &  $0.28 \pm 0.22$ & $0.86 \pm 0.02$ \\
        \hline
        \ce{CatBoost} & $0.71 \pm 0.11$ & $0.17 \pm 0.14$ & $0.21 \pm 0.15$ & $0.16 \pm 0.15$ & $0.9 \pm 0.02$ \\
        \hline
        \ce{TabPFN} & $0.67 \pm 0.11$ & $0.06 \pm 0.06$ & $0.13 \pm 0.16$ & $0.04 \pm 0.04$ & $0.9 \pm 0.02$ \\
        \hline
        \ce{KAN} & $\textbf{0.72} \pm 0.11$ & $\textbf{0.24} \pm 0.09$ & $0.15 \pm 0.06$ & $\textbf{0.54} \pm 0.18$ & $0.73 \pm 0.05$\\
        \hline
        \ce{XGBoost} & $0.69 \pm 0.10$ & $0.18 \pm 0.16$ & $0.22 \pm 0.20$ & $0.17 \pm 0.16$ & $0.89 \pm 0.02$ \\
        \hline
        \multicolumn{6}{c}{Models with TVAE oversampling} \\
        \hline
        \ce{Logistic Regression} & $\textbf{0.68} \pm 0.11$ & $0.00 \pm 0.00$ & $0.00 \pm 0.00$ & $0.00 \pm 0.00$ & $\textbf{0.92} \pm 0.00$ \\
        \hline
        \ce{Random Forest} & $0.67 \pm 0.09$ & $0.15 \pm 0.10$ & $0.18 \pm 0.12$ &  $0.13 \pm 0.09$ & $0.89 \pm 0.01$ \\
        \hline
        \ce{CatBoost} & $0.66 \pm 0.09$ & $0.07 \pm 0.11$ & $0.18 \pm 0.31$ & $0.05 \pm 0.07$ & $0.91 \pm 0.02$ \\
        \hline
        \ce{TabPFN} & $0.61 \pm 0.08$ & $0.01 \pm 0.04$ & $0.10 \pm 0.30$ & $0.01 \pm 0.02$ & $\textbf{0.92} \pm 0.01$ \\
        \hline
        \ce{KAN} & $0.59 \pm 0.07$ & $\textbf{0.18} \pm 0.06$ & $0.12 \pm 0.04$ & $\textbf{0.40} \pm 0.13$ & $0.72 \pm 0.05$ \\
        \hline
        \ce{XGBoost} & $0.63 \pm 0.07$ & $0.1 \pm 0.11$ & $\textbf{0.23} \pm 0.29$ & $0.07 \pm 0.09$ & $0.91 \pm 0.01$ \\
        \hline
        \multicolumn{6}{c}{Models with GAN oversampling} \\
        \hline
        \ce{Logistic Regression} & $0.64 \pm 0.07$ & $0.06 \pm 0.08$ & $0.14 \pm 0.19$ & $0.04 \pm 0.05$ & $\textbf{0.91} \pm 0.01$ \\
        \hline
        \ce{Random Forest} & $0.65 \pm 0.10$ & $\textbf{0.17} \pm 0.11$ & $\textbf{0.23} \pm 0.20$ &  $0.16 \pm 0.10$ & $0.88 \pm 0.03$ \\
        \hline
        \ce{CatBoost} & $\textbf{0.67} \pm 0.09$ & $0.14 \pm 0.14$ & $0.19 \pm 0.21$ & $0.11 \pm 0.11$ & $0.89 \pm 0.02$ \\
        \hline
        \ce{TabPFN} & $0.64 \pm 0.08$ & $0.01 \pm 0.04$ & $0.03 \pm 0.08$ & $0.01 \pm 0.02$ & $\textbf{0.91} \pm 0.01$ \\
        \hline
        \ce{KAN} & $0.63 \pm 0.08$ & $\textbf{0.17} \pm 0.09$ & $0.12 \pm 0.06$ & $\textbf{0.31} \pm 0.17$ & $0.78 \pm 0.04$ \\
        \hline
        \ce{XGBoost} & $0.65 \pm 0.10$ & $0.11 \pm 0.09$ & $0.17 \pm 0.13$ & $0.08 \pm 0.07$ & $0.9 \pm 0.01$ \\
        \hline
        \multicolumn{6}{c}{Models with Gaussian Copula oversampling} \\
        \hline
        \ce{Logistic Regression} & $\textbf{0.7} \pm 0.11$ & $0.00 \pm 0.00$ & $0.00 \pm 0.00$ & $0.00 \pm 0.00$ & $\textbf{0.92} \pm 0.00$ \\
        \hline
        \ce{Random Forest} & $0.67 \pm 0.12$ & $\textbf{0.2} \pm 0.11$ & $\textbf{0.21} \pm 0.11$ &  $0.2 \pm 0.12$ & $0.88 \pm 0.02$ \\
        \hline
        \ce{CatBoost} & $0.67 \pm 0.11$ & $0.07 \pm 0.09$ & $0.1 \pm 0.14$ & $0.05 \pm 0.07$ & $0.9 \pm 0.02$ \\
        \hline
        \ce{TabPFN} & $0.64 \pm 0.11$ & $0.06 \pm 0.08$ & $0.12 \pm 0.16$ & $0.04 \pm 0.05$ & $0.91 \pm 0.01$ \\
        \hline
        \ce{KAN} & $0.64 \pm 0.11$ & $0.19 \pm 0.08$ & $0.12 \pm 0.05$ & $\textbf{0.44} \pm 0.19$ & $0.70 \pm 0.05$\\
        \hline
        \ce{XGBoost} & $0.65 \pm 0.12$ & $0.07 \pm 0.09$ & $0.1 \pm 0.14$ & $0.05 \pm 0.07$ & $0.9 \pm 0.02$ \\
        \hline
        \multicolumn{6}{c}{Models with TabSyn oversampling} \\
        \hline
        \ce{Logistic Regression} & $0.66 \pm 0.09$ & $0.04 \pm 0.06$ & $0.05 \pm 0.07$ & $0.04 \pm 0.05$ & $0.87 \pm 0.02$ \\
        \hline
        \ce{Random Forest} & $\textbf{0.68} \pm 0.09$ & $0.04 \pm 0.06$ & $0.06 \pm 0.09$ &  $0.04 \pm 0.05$ & $\textbf{0.88} \pm 0.02$ \\
        \hline
        \ce{CatBoost} & $0.66 \pm 0.1$ & $0.04 \pm 0.06$ & $0.06 \pm 0.09$ & $0.04 \pm 0.05$ & $\textbf{0.88} \pm 0.02$ \\
        \hline
        \ce{TabPFN} & $0.66 \pm 0.09$ & $0.05 \pm 0.07$ & $0.07 \pm 0.01$ & $0.04 \pm 0.05$ & $\textbf{0.88} \pm 0.02$ \\
        \hline
        \ce{KAN} & $0.66 \pm 0.09$ & $\textbf{0.13} \pm 0.08$ & $\textbf{0.12} \pm 0.07$ & $\textbf{0.16} \pm 0.1$ & $0.85 \pm 0.02$\\
        \hline
        \ce{XGBoost} & $0.67 \pm 0.1$ & $0.05 \pm 0.07$ & $0.06 \pm 0.09$ & $0.04 \pm 0.05$ & $0.87 \pm 0.02$ \\
        \hline
        \multicolumn{6}{c}{Models with Edge cases oversampling} \\
        \hline
        \ce{Logistic Regression} & $0.7 \pm 0.11$ & $0.03 \pm 0.08$ & $0.07 \pm 0.2$ & $0.02 \pm 0.05$ & $\textbf{0.92} \pm 0.01$ \\
        \hline
        \ce{Random Forest} & $0.71 \pm 0.11$ & $0.01 \pm 0.04$ & $0.05 \pm 0.15$ &  $0.01 \pm 0.02$ & $\textbf{0.92} \pm 0$ \\
        \hline
        \ce{CatBoost} & $0.71 \pm 0.1$ & $0.05 \pm 0.07$ & $0.23 \pm 0.33$ & $0.03 \pm 0.04$ & $\textbf{0.92} \pm 0$ \\
        \hline
        \ce{TabPFN} & $0.70 \pm 0.1$ & $0 \pm 0$ & $0 \pm 0$ & $0 \pm 0$ & $\textbf{0.92} \pm 0$ \\
        \hline
        \ce{KAN} & $0.70 \pm 0.09$ & $\textbf{0.18} \pm 0.15$ & $\textbf{0.32} \pm 0.32$ & $\textbf{0.13} \pm 0.11$ & $0.91 \pm 0.02$\\
        \hline
        \ce{XGBoost} & $\textbf{0.72} \pm 0.11$ & $0.03 \pm 0.06$ & $0.15 \pm 0.32$ & $0.02 \pm 0.03$ & $\textbf{0.92} \pm 0$ \\
    \bottomrule
    \end{tabular}
    \label{table:cv_class_metr_full}
\end{table}

\begin{table}[H]
    \caption{Probabilistic metrics on Cross validation}
    \label{validation}
    \centering
    \setlength{\tabcolsep}{10pt}
    \begin{tabular}{llllll}
    \toprule
        {Model} & {Avg Confidence} & {Avg Entropy} & {Brier score} & {ECE}\\
        \midrule
        \multicolumn{4}{c}{Models without oversampling} \\
        \hline
        \ce{Logistic Regression} & $0.92 \pm 0$ & $0.27 \pm 0$ & $0.07 \pm 0$ & $0.03 \pm 0.03$ \\
        \hline
        \ce{Random Forest} & $0.92 \pm 0.01$ & $0.26 \pm 0.01$ & $0.07 \pm 0.01$ & $0.03 \pm 0.02$ \\
        \hline
        \ce{CatBoost} & $0.92 \pm 0.01$ & $0.26 \pm 0.01$ & $0.07 \pm 0.01$ & $0.02 \pm 0.02$ \\
        \hline
        \ce{TabPFN} & $0.93 \pm 0.01$ & $0.22 \pm 0.01$ & $0.07 \pm 0.01$ & $0.03 \pm 0.01$\\
        \hline
        \ce{KAN} & $0.92 \pm 0.01$ & $0.26 \pm 0.01$ & $0.07 \pm 0.01$ & $0.03 \pm 0.02$ \\
        \hline
        \ce{XGBoost} & $0.92 \pm 0.01$ & $0.26 \pm 0.01$ & $0.07 \pm 0.01$ & $0.03 \pm 0.02$ \\
        \hline
        \multicolumn{4}{c}{Models with ARF oversampling} \\
        \hline
        \ce{Logistic Regression} & $0.70 \pm 0.01$ & $0.6 \pm 0.01$ & $0.12 \pm 0.01$ & $0.23 \pm 0.01$ \\
        \hline
        \ce{Random Forest} & $0.82 \pm 0.01$ & $0.42 \pm 0.02$ & $0.10 \pm 0.01$ & $0.14 \pm 0.02$ \\
        \hline
        \ce{CatBoost} & $0.88 \pm 0.01$ & $0.32 \pm 0.01$ & $0.09 \pm 0.02$ & $0.07 \pm 0.01$ \\
        \hline
        \ce{TabPFN} & $0.92 \pm 0$ & $0.24 \pm 0.01$ & $0.08 \pm 0.01$ & $0.06 \pm 0.02$\\
        \hline
        \ce{KAN} & $0.67 \pm 0.01$ & $0.60 \pm 0.01$ & $0.19 \pm 0.02$ & $0.33 \pm 0.02$ \\
        \hline
        \ce{XGBoost} & $0.87 \pm 0.01$ & $0.33 \pm 0.01$ & $0.09 \pm 0.02$ & $0.09 \pm 0.02$ \\
        \hline
        \multicolumn{4}{c}{Models with TVAE oversampling} \\
        \hline
        \ce{Logistic Regression} & $0.71 \pm 0.02$ & $0.59 \pm 0.02$ & $0.11 \pm 0.01$ & $0.21 \pm 0.02$ \\
        \hline
        \ce{Random Forest} & $0.87 \pm 0.01$ & $0.33 \pm 0.02$ & $0.09 \pm 0.01$ & $0.08 \pm 0.02$ \\
        \hline
        \ce{CatBoost} & $0.90 \pm 0.01$ & $0.29 \pm 0.01$ & $0.08 \pm 0.01$ & $0.05 \pm 0.01$ \\
        \hline
        \ce{TabPFN} & $0.92 \pm 0$ & $0.24 \pm 0.01$ & $0.07 \pm 0.01$ & $0.04 \pm 0.01$ \\
        \hline
        \ce{KAN} & $0.71 \pm 0.02$ & $0.54 \pm 0.02$ & $0.19 \pm 0.02$ & $0.29 \pm 0.03$ \\
        \hline
        \ce{XGBoost} & $0.89 \pm 0.01$ & $0.3 \pm 0.02$ & $0.08 \pm 0.01$ & $0.06 \pm 0.02$ \\
        \hline
        \multicolumn{4}{c}{Models with GAN oversampling} \\
        \hline
        \ce{Logistic Regression} & $0.72 \pm 0.01$ & $0.58 \pm 0.01$ & $0.12 \pm 0.01$ & $0.21 \pm 0.01$ \\
        \hline
        \ce{Random Forest} & $0.85 \pm 0.01$ & $0.37 \pm 0.02$ & $0.1 \pm 0.02$ & $0.1 \pm 0.02$ \\
        \hline
        \ce{CatBoost} & $0.9 \pm 0.01$ & $0.29 \pm 0.01$ & $0.08 \pm 0.01$ & $0.05 \pm 0.01$ \\
        \hline
        \ce{TabPFN} & $0.93 \pm 0$ & $0.23 \pm 0.01$ & $0.08 \pm 0.01$ & $0.05 \pm 0.02$ \\
        \hline
        \ce{KAN} & $0.76 \pm 0.03$ & $0.5 \pm 0.04$ & $0.16 \pm 0.02$ & $0.23 \pm 0.05$\\
        \hline
        \ce{XGBoost} & $0.87 \pm 0.01$ & $0.34 \pm 0.02$ & $0.09 \pm 0.01$ & $0.07 \pm 0.01$ \\
        \hline
        \multicolumn{4}{c}{Models with Gaussian Copula oversampling} \\
        \hline
        \ce{Logistic Regression} & $0.69 \pm 0$ & $0.61 \pm 0$ & $0.12 \pm 0$ & $0.23 \pm 0.01$ \\
        \hline
        \ce{Random Forest} & $0.81 \pm 0.01$ & $0.43 \pm 0.01$ & $0.1 \pm 0.01$ & $0.14 \pm 0.01$ \\
        \hline
        \ce{CatBoost} & $0.90 \pm 0$ & $0.3 \pm 0$ & $0.09 \pm 0.01$ & $0.06 \pm 0.02$ \\
        \hline
        \ce{TabPFN} & $0.93 \pm 0$ & $0.23 \pm 0.01$ & $0.08 \pm 0.01$ & $0.05 \pm 0.02$ \\
        \hline
        \ce{KAN} & $0.63 \pm 0.01$ & $0.64 \pm 0.01$ & $0.2 \pm 0.01$ & $0.35 \pm 0.01$\\
        \hline
        \ce{XGBoost} & $0.89 \pm 0.01$ & $0.32 \pm 0.01$ & $0.09 \pm 0.01$ & $0.06 \pm 0.02$ \\
        \hline
        \multicolumn{4}{c}{Models with TabSyn oversampling} \\
        \hline
        \ce{Logistic Regression} & $0.9 \pm 0$ & $0.29 \pm 0.01$ & $0.09 \pm 0.01$ & $0.09 \pm 0.03$ \\
        \hline
        \ce{Random Forest} & $0.91 \pm 0$ & $0.28 \pm 0.01$ & $0.1 \pm 0.01$ & $0.08 \pm 0.03$ \\
        \hline
        \ce{CatBoost} & $0.91 \pm 0$ & $0.29 \pm 0.01$ & $0.1 \pm 0.01$ & $0.07 \pm 0.03$ \\
        \hline
        \ce{TabPFN} & $0.92 \pm 0.01$ & $0.23 \pm 0.01$ & $0.1 \pm 0.01$ & $0.07 \pm 0.03$ \\
        \hline
        \ce{KAN} & $0.87 \pm 0.01$ & $0.33 \pm 0.01$ & $0.12 \pm 0.02$ & $0.12 \pm 0.02$\\
        \hline
        \ce{XGBoost} & $0.91 \pm 0$ & $0.28 \pm 0.01$ & $0.1 \pm 0.01$ & $0.07 \pm 0.03$ \\
        \hline
        \multicolumn{4}{c}{Models with Edge cases oversampling} \\
        \hline
        \ce{Logistic Regression} & $0.91 \pm 0$ & $0.3 \pm 0.01$ & $0.07 \pm 0.01$ & $0.04 \pm 0.02$ \\
        \hline
        \ce{Random Forest} & $0.92 \pm 0$ & $0.26 \pm 0.01$ & $0.07 \pm 0.01$ & $0.03 \pm 0.02$ \\
        \hline
        \ce{CatBoost} & $0.92 \pm 0.01$ & $0.26 \pm 0.01$ & $0.07 \pm 0.01$ & $0.03 \pm 0.02$ \\
        \hline
        \ce{TabPFN} & $0.94 \pm 0$ & $0.2 \pm 0.01$ & $0.07 \pm 0.01$ & $0.04 \pm 0.01$ \\
        \hline
        \ce{KAN} & $0.88 \pm 0.01$ & $0.31 \pm 0.01$ & $0.08 \pm 0.01$ & $0.06 \pm 0.01$\\
        \hline
        \ce{XGBoost} & $0.92 \pm 0$ & $0.27 \pm 0.01$ & $0.07 \pm 0$ & $0.03 \pm 0.02$ \\
        
    \bottomrule
    \end{tabular}
    \label{table:cv_prob_metr_full}
\end{table}

\textbf{Feature descriptors}: Age was defined as the age of the patient in years at the time of the index procedure. Anemia was defined as baseline hematocrit <39\% for men and <36\% for women. Ejection fraction was defined as the left ventricular ejection fraction (LVEF) measured by echocardiography prior to the intervention. Cerebrovascular disease was defined as a history of obstructive atherosclerotic lesion in the carotid or vertebral arteries, a previous or planned intervention on cerebrovascular vessels, or a documented history of stroke. Chronic kidney disease (CKD) was defined as an estimated glomerular filtration rate (eGFR) <60 ml/min/1.73 $m^2$ persisting for three months or longer, irrespective of the underlying cause. Peripheral artery disease was defined as a history of atherosclerotic obstructive lesions in the aorta or limb arteries, previous or planned intervention on peripheral vessels, or prior amputation for arterial disease. Aortic stenosis was defined as aortic valve area <1.0 $cm^2$ or mean transvalvular gradient $\geq$ 40 mmHg, diagnosed by echocardiography. Single vessel disease was defined as the presence of significant obstructive stenosis ($\geq$ 70\%) in only one major epicardial coronary artery. Coronary calcium was defined as the angiographic presence of any calcification within the target vessel. Stent type (Calypso, Synergy, Xience) refers to the specific drug-eluting stent platform implanted during the procedure. Medina side classification was defined according to the Medina system for coronary bifurcations, indicating side branch involvement. Atrial fibrillation was defined as a documented history of atrial fibrillation or flutter on medical records or ECG. The DEFINITION score was applied as a bifurcation lesion stratification system, with major and minor criteria as described previously. History of cancer was defined as a documented diagnosis of any malignant neoplasm prior to the index procedure. Stent diameter was defined as the nominal diameter of the implanted stent in millimeters. Stent length was defined as the nominal length of the implanted stent in millimeters. Ad-hoc PCI was defined as PCI performed during the same procedure as the diagnostic coronary angiography. Previous PCI was defined as any prior percutaneous coronary intervention before the index procedure. CTO bifurcation was defined as the presence of a bifurcation lesion within or adjacent to a chronic total occlusion (CTO).

\begin{table}[H]
    \caption{Distribution of features between samples for training, testing and edge cases.}
    \centering
    \begin{tabular}{llllll}
    \toprule

        {Feature} & {Train} & {Test} & {Edge cases} & {External validation}\\
        \midrule
        \ce{Age} & $63.9\pm9.85$ & $63.5\pm9.3$ & $92.5 \pm 4.33$ & $61.6 \pm 11.13$\\
        \hline
        \ce{Anemia} & 5\% & 4.4\% & 90\% & 18\%\\
        \hline
        \ce{Ejection Fraction} & $56.2\pm10.6$ &  $55.9\pm10.5$ &  $22.5\pm4.33$ & $45.2 \pm 6.31$\\
        \hline
        \ce{Cerebrovascular Disease} & 12.3\% & 13.4\% & 90\% & 3.8\%\\
        \hline
        \ce{CKD} & $75.3\pm16.8$ & $76\pm16$ & $30.0\pm8.66$ & $90.2 \pm 29.6$\\
        \hline
        \ce{Peripheral Artery Disease } & 7.8\% & 9.5\% & 85\% & 0\% \\
        \hline
        \ce{Aortic Stenosis} & 2.4\% & 1.7\% & 80\% & 0\% \\
        \hline
        \ce{Single Vessel disease} & 46.5\% & 44\% & 90\% & 58.2\%\\
        \hline
        \ce{Coronary calcium} & 20.9\% & 19.8\% & 90\% & 16.5\%\\
        \hline
        \ce{Stent type - Calypso} & 36.2\% & 38.4\% & 70\% & 0\%\\
        \hline
        \ce{Medina side} & 33.6\% & 35\% & 80\% & 0\%\\
        \hline
        \ce{Atrial Fibrillation} & 14.3\% & 11.7\% & 80\% & 8\% \\
        \hline
        \ce{DEFINITION score} & 0.6\% & 0.5\% & 90\% & 0\% \\
        \hline
        \ce{History of cancer} & 5.1\% & 4.2\% & 60\% & 0\% \\
        \hline
        \ce{Stent type - Synergy} & 1\% & 1\% & 70\% & 0\% \\
        \hline
        \ce{Stent diameter} &  $3.25\pm0.513$ & $3.22\pm0.511$ & $2.375\pm0.22$ & $3.33 \pm 0.44$\\
        \hline
        \ce{Stent length} & $24.5\pm8.34$ & $23.4\pm8.56$ & $33.0\pm2.89$ & $20.65\pm5.16$\\
        \hline
        \ce{Ad-hoc PCI} & 40.7\% & 38.1\% & 80\% & 100\%\\
        \hline
        \ce{Previous PCI} & 41.3\% & 42.3\% & 80\% & 3.8\% \\
        \hline
        \ce{Stent type - Xience} & 12.0\% & 8.3\% & 70\% & 0\% \\
        \hline
        \ce{CTO bifurcation} & 8.1\% & 8.8\% & 80\% & 0\%\\
    \bottomrule
    \end{tabular}
    \label{table:features_dataset}
\end{table}

\textbf{Metrics}

The performance of the models was evaluated using commonly applied classification metrics: area under the receiver operating characteristic curve (AUC-ROC), F1-score, precision, recall, and accuracy. In addition, to assess the reliability of predicted probabilities and the confidence of the models, we employed several probabilistic metrics: average confidence, average entropy, the Brier score, and expected calibration error (ECE).

\begin{equation}
\text{Accuracy} = \frac{TP + TN}{TP + TN + FP + FN}
\end{equation}

\begin{equation}
\text{Precision} = \frac{TP}{TP + FP}
\end{equation}

\begin{equation}
\text{Recall} = \frac{TP}{TP + FN}
\end{equation}

\begin{equation}
F1 = 2 \cdot \frac{\text{Precision} \cdot \text{Recall}}{\text{Precision} + \text{Recall}}
\end{equation}

Area under the Receiver Operating Characteristic curve, computed as:

\begin{equation}
\text{AUC} = \int_0^1 TPR(FPR) \, dFPR
\end{equation}

where $TPR$ and $FPR$ denote true positive rate and false positive rate respectively.

\begingroup
\raggedbottom
\setlength{\abovedisplayskip}{6pt}
\setlength{\belowdisplayskip}{6pt}
\setlength{\abovedisplayshortskip}{6pt}
\setlength{\belowdisplayshortskip}{6pt}

\begin{gather}
\text{AvgConf} = \frac{1}{N}\sum_{i=1}^{N}\max_{c} p_i(c) \\[4pt]
\text{AvgEntropy} = -\frac{1}{N}\sum_{i=1}^{N}\sum_{c=1}^{C} p_i(c)\log p_i(c) \\[4pt]
\text{Brier} = \frac{1}{N}\sum_{i=1}^{N}\sum_{c=1}^{C}\big(p_i(c)-y_i(c)\big)^2 \\[4pt]
\text{ECE} = \sum_{m=1}^{M}\frac{|B_m|}{N}\,\big|\text{acc}(B_m)-\text{conf}(B_m)\big|
\end{gather}
\endgroup

\textbf{Probabilistic measures}
In addition to standard classification metrics, probabilistic measures were employed to evaluate the reliability of predicted probabilities, which is particularly important in the presence of class imbalance. Average Confidence quantifies the mean predicted confidence of the model, reflecting whether the classifier tends to be overconfident in its decisions. Average Entropy measures the uncertainty of the predicted probability distributions; in imbalanced settings, excessively low entropy may indicate that the model consistently favors the majority class without adequately representing uncertainty in difficult cases.

The Brier score captures the mean squared error between predicted probabilities and the true outcomes, thereby providing an overall measure of calibration. A low Brier score indicates that predicted probabilities are well aligned with observed frequencies, which is essential when models are used to estimate risk rather than to provide only binary classifications. Similarly, the Expected Calibration Error (ECE) quantifies the discrepancy between predicted confidence and actual accuracy across probability bins. High ECE values reveal that predicted probabilities may be systematically miscalibrated, for example, when the model outputs overly high probabilities for the majority class.

\begin{figure}[H]
    \centering
    \begin{subfigure}{0.48\textwidth}
        \centering
        \includegraphics[width=\linewidth]{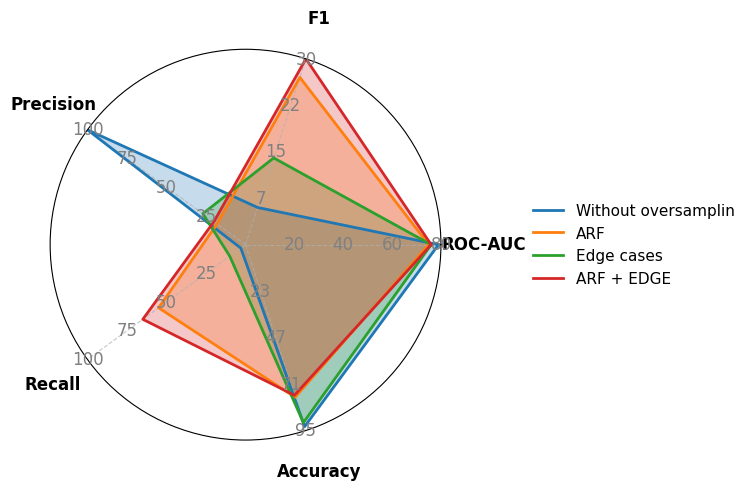}
        \caption{Radar plot of classification metrics for KAN model.}
        \label{fig:radar_kan}
    \end{subfigure}
    \hfill
    \begin{subfigure}{0.48\textwidth}
        \centering
        \includegraphics[width=\linewidth]{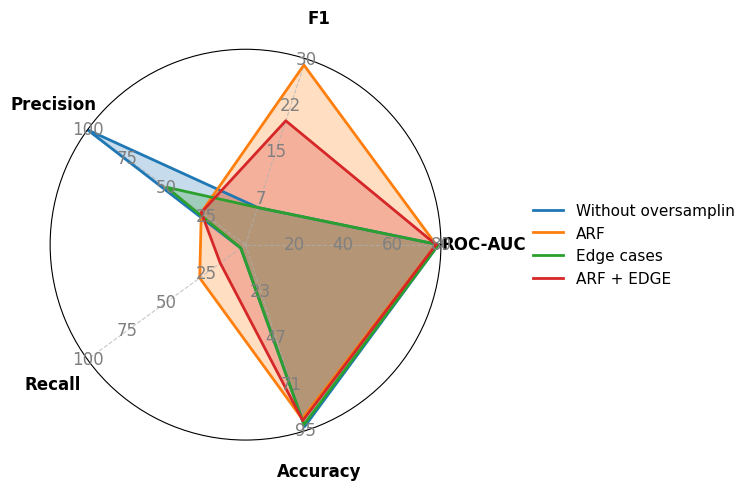}
        \caption{Radar plot of classification metrics for XGBoost.}
        \label{fig:radar_xgb}
    \end{subfigure}
    
    \caption{Radar plots of classification metrics on test (Without classification, ARF, Edge cases, ARF + Edge cases).}
    \label{fig:Radar_plots}
\end{figure}

\textbf{Radar plots}
To demonstrate the differences in metrics across various approaches to synthetic data generation, we employed radar plots in figure \ref{fig:Radar_plots} for the results of the KAN and XGBoost models. The plots reveal that, without synthetic data, both models achieve a very high Precision score of 1, while Recall and F1 score remain at very low levels. The application of ARF and ARF with edge cases improves Recall and F1, albeit at the expense of Precision. In the radar plot of KAN results, it can be observed that the combination of ARF and edge cases further enhances the metrics compared to their individual application. In contrast, for XGBoost, the combination of synthetic data methods leads to a slight decline in performance metrics.

\begin{figure}[H]
    \centering
    \begin{subfigure}{0.45\textwidth}
        \centering
        \includegraphics[width=\linewidth]{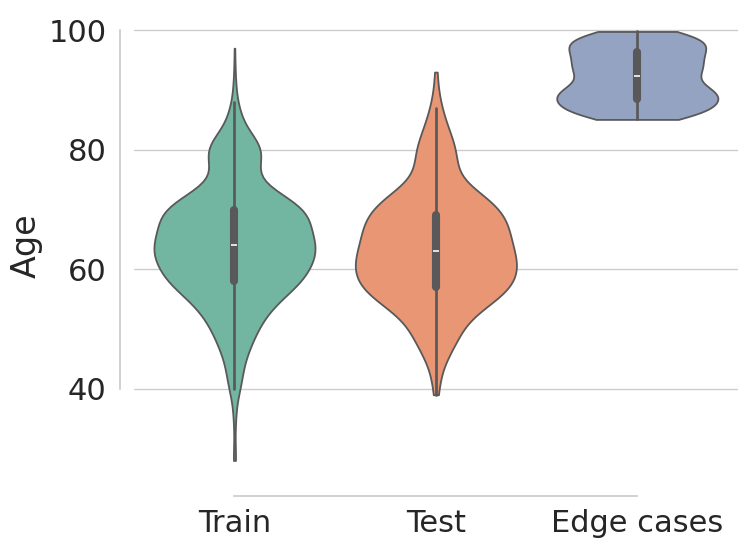}
        \caption{Distribution of age across datasets.}
        \label{fig:age}
    \end{subfigure}
    \hfill
    \begin{subfigure}{0.45\textwidth}
        \centering
        \includegraphics[width=\linewidth]{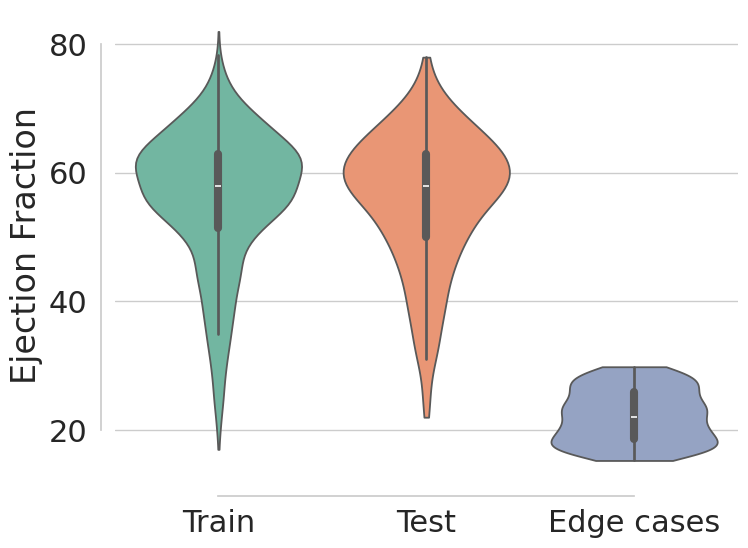}
        \caption{Distribution of ejection fraction across datasets.}
        \label{fig:ef}
    \end{subfigure}
    
    \vspace{0.5cm}
    
    \begin{subfigure}{0.45\textwidth}
        \centering
        \includegraphics[width=\linewidth]{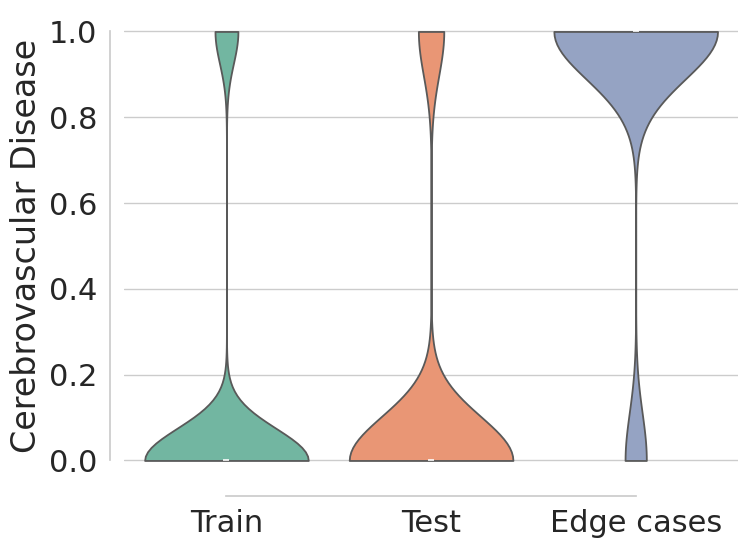}
        \caption{Distribution of cerebrovascular disease across datasets.}
        \label{fig:cd}
    \end{subfigure}
    \hfill
    \begin{subfigure}{0.45\textwidth}
        \centering
        \includegraphics[width=\linewidth]{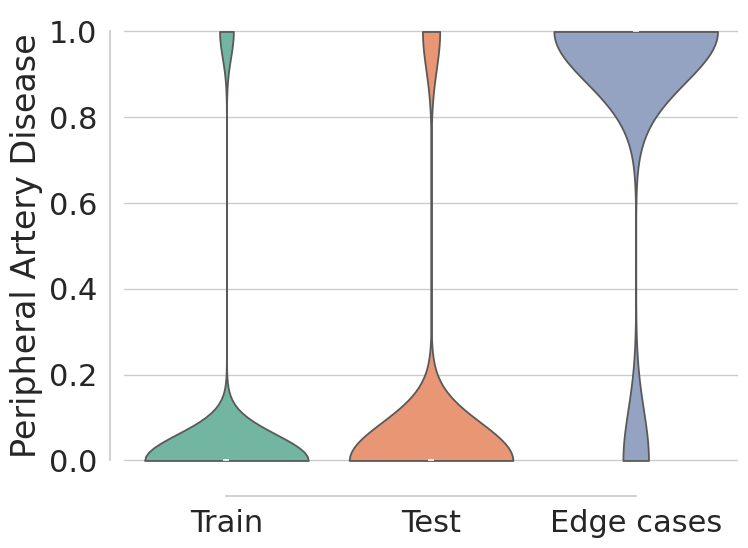}
        \caption{Distribution of peripheral artery disease across datasets.}
        \label{fig:pad}
    \end{subfigure}

    \caption{Distributions of clinical characteristics across datasets.}
    \label{fig:all_distributions}
\end{figure}

To illustrate the differences (see figure ~\ref{fig:all_distributions}) in the distribution of features that contribute most to model classification, we employed violin plots for the training, test, and edge-case cohorts. The features with the highest contribution to classification performance include age, ejection fraction (EF), cerebrovascular disease, and peripheral artery disease. As shown in the plots, the distributions of features in the training and test sets are largely consistent. The median values of age and ejection fraction are nearly identical across models. Similarly, for cerebrovascular disease and peripheral artery disease, the majority of patients in both the training and test sets do not present these conditions. In contrast, by construction, the edge-case cohort comprises predominantly older patients with reduced ejection fraction and a comparatively higher prevalence of cerebrovascular and peripheral artery disease.

\end{document}